\documentclass[final]{cvpr}

\usepackage{times}
\usepackage{epsfig}
\usepackage{graphicx}
\usepackage{amsmath}
\usepackage{amssymb}
\usepackage{bm}
\usepackage{algorithm}
\usepackage{algorithmic}
\usepackage{subcaption}
\usepackage{booktabs}

\usepackage[pagebackref=true,breaklinks=true,colorlinks,bookmarks=false]{hyperref}

\usepackage[toc,page]{appendix}

\def\cvprPaperID{10043}
\def\confYear{CVPR 2021}

\begin{document} 

\title{Post-hoc Uncertainty Calibration for Domain Drift Scenarios}

\author{
    Christian Tomani\textsuperscript{\rm 1}, Sebastian Gruber\textsuperscript{\rm 3,4,5,*}, Muhammed Ebrar Erdem\textsuperscript{\rm 2},\\ Daniel Cremers\textsuperscript{\rm 1}, Florian Buettner\textsuperscript{\rm 2,3,4,5,*}\\
    \textsuperscript{\rm 1}Technical University of Munich\\ \textsuperscript{\rm 2}Siemens AG \hspace{10pt}
    \textsuperscript{\rm 3} German Cancer Consortium\\
    \textsuperscript{\rm 4} German Cancer Research Center Heidelberg
    \hspace{10pt}
    \textsuperscript{\rm 5} Frankfurt University\\ 
    \tt\small \{christian.tomani, cremers\}@tum.de,\\
    \tt\small \{sebastian.gruber, muhammed.erdem, buettner.florian\}@siemens.com

}

\maketitle
\begin{abstract} 
We address the problem of uncertainty calibration. While standard deep neural networks typically yield uncalibrated predictions, calibrated confidence scores that are representative of the true likelihood of a prediction can be achieved using post-hoc calibration methods. However, to date, the focus of these approaches has been on in-domain calibration. Our contribution is two-fold. First, we show that existing post-hoc calibration methods yield highly over-confident predictions under domain shift. Second, we introduce a simple strategy where perturbations are applied to samples in the validation set before performing the post-hoc calibration step. In extensive experiments, we demonstrate that this perturbation step results in substantially better calibration under domain shift on a wide range of architectures and modelling tasks.

\end{abstract} 

\section{Introduction}\label{sec:intro}
\subsection{Towards calibrated classifiers}
\footnotetext{\textsuperscript{\rm * }Work done for Siemens AG}

\begin{figure}[ht!]
	\centering
	\includegraphics[width=0.45\textwidth]{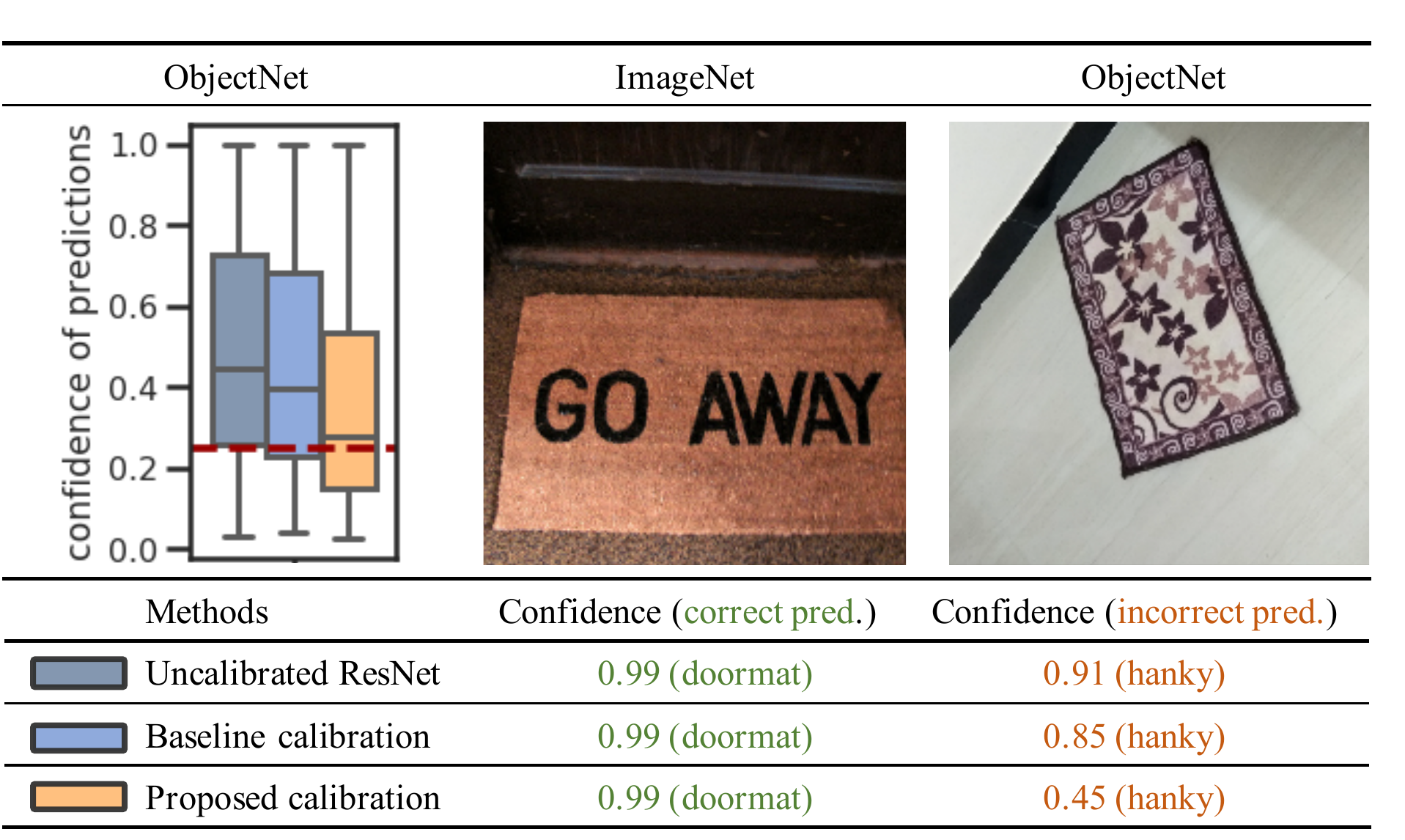}
	\caption{Re-calibrated neural networks make over-confident predictions under domain drift. Left: Only for our approach confidence scores match model accuracy (dashed red line) across all predictions; all other approaches are over-confident under domain shift. Middle: For Imagenet, all models make a correct prediction with high confidence (left). Right: Under domain drift (different viewpoint; Objectnet) all models make wrong predictions, but only our approach has a low confidence score reflecting model uncertainty. }
	\label{fig:teaset}
\end{figure}

Due to their high predictive power, deep neural networks are increasingly being used as part of decision  making systems in real world applications. However, such systems require not only high accuracy, but also reliable and calibrated uncertainty estimates. A classifier is {\em calibrated}, if the confidence of predictions matches the probability of being correct for all confidence levels \cite{guo_calibration_2017}. Especially in safety critical applications in medicine where average case performance is insufficient, but also in dynamically changing environments in industry, practitioners need to have access to reliable predictive uncertainty during the entire life-cycle of the model. This means confidence scores (or predictive uncertainty) should be well calibrated not only for in-domain predictions, but also under gradual domain drift where the distribution of the input samples gradually changes from in-domain to truly out-of-distribution (OOD). Such domain drift scenarios commonly include changes in object backgrounds, rotations, and imaging viewpoints \cite{barbu2019objectnet}; Fig. \ref{fig:teaset}.

Since deep neural networks typically only yield uncalibrated confidence scores, a variety of different post-hoc calibration approaches have been proposed \cite{Platt99probabilisticoutputs,guo_calibration_2017,zadrozny2002transforming,zadrozny2001obtaining,zhang2020mix}. These methods use the validation set  to transform predictions returned by a trained neural network such that in-domain predictions are well calibrated. Such post-hoc uncertainty calibration approaches are particularly appealing since costly training of intrinsically uncertainty-aware neural networks can be avoided.\\
Current efforts to systematically quantify the quality of predictive uncertainties have focused on assessing model calibration for in-domain predictions. Here, post-processing in form of temperature scaling has shown great promise and Guo \emph{et~al.}~\cite{guo_calibration_2017} illustrated that this approach yields well calibrated predictions for a wide range of model architectures. More recently, more complex combinations of parametric and non-parametric methods have been proposed \cite{zhang2020mix}. However, little attention has been paid to uncertainty calibration under domain drift and no comprehensive analysis of the performance of post-hoc uncertainty calibration methods under domain drift exists.

\subsection{Contribution}

In this work we focus on the task of post-hoc uncertainty calibration under domain drift scenarios and make the following contributions:
\begin{itemize}
    \item We first show that neural networks yield overconfident predictions  under domain shift even after re-calibration using existing post-hoc calibrators.
    \item We generalise existing post-hoc calibration methods by transforming the validation set before performing the post-hoc calibration step.
    \item We demonstrate that our approach results in substantially better calibration under domain shift on a wide range of architectures and image data sets.
\end{itemize}
In addition to the contributions above, our code is made available at \url{https://github.com/tochris/calibration-domain-drift}.

\section{Related work}
In this section, we review existing approaches towards neural networks with calibrated predictive uncertainty. The focus of this work is on post-hoc calibration methods, which we review in detail. These approaches can broadly be divided into 2 categories: accuracy-preserving methods, where the ranking of confidence scores across classes remain unchanged and those where the ranking, and thus accuracy, can change.  Other related work includes intrinsically uncertainty-aware neural networks and out-of-distribution detection methods. 

\subsection{Post-processing methods} A popular approach towards well-calibrated predictions are post-hoc calibration methods where a validation set, drawn from the generative distribution of the training data $\pi(X,Y)$, is used to rescale the outputs returned by a trained neural network such that in-domain predictions are well calibrated. A variety of parametric, as well as non-parametric methods exist. We first review non-parametric methods that do not preserve accuracy. A simple but popular non-parametric post-processing approach is histogram binning \cite{zadrozny2001obtaining}. In brief, all uncalibrated confidence scores $\hat{P}_l$  are partitioned into $M$ bins (with borders typically chosen such that either all bins are of equal size or contain the same number of samples). Next, a calibrated score $Q_m$ that is determined by optimizing a bin-wise squared loss on the validation set, is assigned to each bin. For each new prediction, the uncalibrated confidence score  $\hat{P}_{pr}$ is then replaced by the calibrated score associated with the bin $\hat{P}_{pr}$ falls into. Popular extensions to histogram binning are isotonic regression \cite{zadrozny2002transforming} and Bayesian Binning into Quantiles (BBQ) \cite{naeini2015obtaining}. For isotonic regression (IR), uncalibrated confidence scores are divided into $M$ intervals and a piecewise constant function $f$ is fitted on the validation set to transform uncalibrated outputs to calibrated scores. BBQ is a Bayesian generalisation of histogram binning based on the concept of Bayesian model averaging.\\
In addition to these non-parametric approaches, also parametric alternatives for post-processing confidence scores exist. Platt scaling \cite{Platt99probabilisticoutputs} is an approach for transforming the non-probabilistic outputs (logits) $z_i \in \mathbb{R}$ of a binary classifier to calibrated confidence scores. More specifically, the logits are transformed to calibrated confidence scores $\hat{Q}_i$ using logistic regression $\hat{Q}_i= \sigma(a z_i + b)$, where $\sigma$ is the sigmoid function and the two parameters $a$ and $b$ are fitted by optimising the negative log-likelihood of the validation set.\\
Guo \emph{et~al.}~\cite{guo_calibration_2017} have proposed Temperature Scaling (TS), a simple generalisation of Platt scaling to the multi-class case, where a single scalar parameter $T$ is used to re-scale the logits of a trained neural network. In the case of $C$-class classification, the logits are a $C$-dimensional vector $\mathbf{z}_i \in \mathbb{R}^C$, which are typically transformed into confidence scores $\hat{P}_i$ using the softmax function $\sigma_{SM}$. For temperature scaling, logits are rescaled with temperature $T$ and transformed into calibrated confidence scores $\hat{Q}_i$ using the softmax function as 
\begin{equation}
\hat{Q}_i = \max_c \sigma_{SM}(\mathbf{z}_i/T)^{(c)}    
\end{equation}\label{eq:ts}
$T$ is learned by minimizing the negative log likelihood of the validation set. In contrast to the non-parametric methods introduced above or other multi-class generalisations of Platt scaling such as vector scaling or matrix scaling, Temperature Scaling has the advantage that it does not change the accuracy of the trained neural network. Since re-scaling  does not affect the ranking of the logits, also the maximum of the softmax function remains unchanged.\\
More recently, Zhang \emph{et~al.}~ \cite{zhang2020mix} have proposed to combine to combine parametric methods with non-parametric methods, in particular they suggest it can be beneficial to perform IR after a TS step (TS-IR). In addition, they introduced an accuracy-preserving version of IR, termed IRM and an ensemble version of temperature scaling, called ETS.\\
 
\subsection{Intrinsically uncertainty-aware neural networks}
A variety of approaches towards intrinsically uncertainty-aware neural networks exist, including probabilistic and non-probabilistic approaches. In recent years, a lot of research effort has been put into training Bayesian neural networks. Since exact inference is untractable, a range of approaches for approximate inference has been proposed,  \cite{gal2016dropout, wen2018flipout}. A popular non-probabilistic alternative is Deep Ensembles \cite{lakshminarayanan_simple_2017}, where an ensemble of networks is trained using adversarial examples, yielding smooth predictions with meaningful predictive uncertainty. Other non-probabilistic alternatives include, e.g., \cite{sensoy_evidential_2018, tomani2021falcon}.\\
While a comprehensive analysis of the performance of post-hoc calibration methods is missing for domain-drift scenarios, recently Ovadia \emph{et~al.}~\cite{snoek2019can} have presented a first comprehensive evaluation of calibration under domain drift for intrinsically uncertainty-aware neural networks and have shown that the quality of predictive  uncertainties, i.e. model calibration, decreases with increasing dataset shift, regardless of method.
\subsection{Other related work}
Orthogonal approaches have been proposed where trust scores and other measures for out-of-distribution detection are derived to detect truly OOD samples, often also based on trained networks and with access to a known OOD set \cite{liang_enhancing_2018,jiang2018trust,papernot2018deep}. However, rather than only detecting truly OOD samples, in this work, we are interested in calibrated confidence scores matching model accuracy at all stages of domain drift, from in-domain samples to truly OOD samples.\\

\section{Problem setup and definitions}

Let $X \in \mathbb{R}^D$ and $Y \in \{1,\dots, C\}$ be random variables that denote the $D$-dimensional input and labels in a classification task with $C$ classes with a ground truth joint distribution $\pi(X, Y ) =
\pi(Y |X)\pi(X)$. The dataset $\mathcal{D}$ consists of $N$ i.i.d.samples $D = \{(X_n, Y_n)\}_{n=1}^N$ drawn from  $\pi(X, Y )$. Let $h(X) = (\hat{Y},\hat{P})$ be the output of a neural network classifier $h$ predicting a class $\hat{Y}$ and associated confidence $\hat{P}$ based on $X$.  Here, we are interested in the quality of predictive uncertainty (i.e. confidence scores $\hat{P}$) not only on test data from the generative distribution of the training data $\mathcal{D}$,  $\pi(X, Y )$, but also under dataset shift, that is test data from a distribution $\rho(X,Y) \neq \pi(X,Y)$. More specifically, we investigate domain drift scenarios where the distribution of samples seen by a model gradually moves away from the training distribution  $\pi(X, Y )$ (in an unknown fashion) until it reaches truly OOD levels.\\
We assess the quality of the confidence scores using the notion of \textit{calibration}. We define perfect calibration such that accuracy and confidence match for all confidence levels:
\begin{eqnarray}
\mathop{\mathbb{P}}(\hat{Y}=Y| \hat{P}=p) = p, \; \; \forall p \in [0,1]
\end{eqnarray}\label{eq:cali}
This directly leads to a definition of miss-calibration as the difference in expectation between confidence and accuracy:
Based on equation \ref{eq:cali} it is straight-forward to define miss-calibration as the difference in expectation between confidence and accuracy:
\begin{eqnarray}
\mathop{\mathbb{E}}_{\hat{P}}\left[\big\lvert\mathop{\mathbb{P}}(\hat{Y}=Y| \hat{P}=p) - p\big\rvert \right]
\label{eq:miscal}
\end{eqnarray}

The expected calibration error (ECE) \cite{naeini2015obtaining} is a scalar summary measure estimating miss-calibration by approximating equation \ref{eq:miscal} based on predictions, confidence scores and ground truth labels $\{(Y_l,\hat{Y}_l,\hat{P}_l)\}_{l=1}^L$ of a finite number of $L$ samples. ECE is computed by first partitioning all $L$ confidence scores $\hat{P}_l$ into $M$ equally sized bins of size $1/M$ and computing accuracy and average confidence of each bin. Let $B_m$ be the set of indices of samples whose confidence falls into its associated interval $I_m = \left(\frac{m-1}{M} ,\frac{m}{M}\right]$. $\mathrm{conf}(B_m) = 1/|B_m|\sum_{i\in B_m}\hat{P}_i$ and $\mathrm{acc}(B_m) = 1/|B_m|\sum_{i\in B_m} \mathbf{1}(\hat{Y}_i = Y_i)$ are the average confidence and accuracy associated with $B_m$, respectively. The ECE is then computed as
\begin{eqnarray}
\mathrm{ECE} = \sum_{m=1}^M \frac{\lvert B_m\rvert}{n}\big\lvert \mathrm{acc}(B_m) - \mathrm{conf}(B_m)\big\rvert
\end{eqnarray}\label{eq:ece}
It can be shown that ECE is directly connected to miss-calibration, as ECE using $M$ bins converges to the $M$-term Riemann-Stieltjes sum of eq. \ref{eq:miscal} \cite{guo_calibration_2017}.\\

\section{Uncertainty calibration under domain drift}

\subsection{Baseline methods and experimental setup}

We assess the performance of the following post-hoc uncertainty calibration methods in domain drift scenarios:
\begin{itemize}
\item Base: Uncalibrated baseline model
\item Temperature scaling (TS) \cite{guo_calibration_2017} and Ensemble Temperature Scaling (ETS) \cite{zhang2020mix}
\item Isotonic Regression (IR) \cite{zadrozny2002transforming}
\item Accuracy preserving version of Isotonic Regression (IRM) \cite{zhang2020mix}
\item Composite model combining Temperature Scaling and Isotonic Regression (TS-IR) \cite{zhang2020mix}
\end{itemize}
We quantify calibration under domain shift for 28 distinct perturbation types not seen during training, including 9 affine transformations, 19 image perturbations introduced by \cite{hendrycks2019benchmarking} and a dedicated bias-controlled dataset \cite{barbu2019objectnet}. Each perturbation strategy mimics a scenario where the data of a deployed model encounters stems from a distribution that gradually shifts away from the training distribution in a different manner.  For each model and each perturbation, we compute the micro-averaged ECE by first perturbing each sample in the test set at 10 different levels  and then calculating the overall ECE across all samples; we denote relative perturbation strength as epsilon. A common manifestation of dataset shift in real-world applications is a change in object backgrounds, rotations, and imaging viewpoints. In order to quantify the expected calibration error under those scenarios, we use Objectnet, a recently proposed large-scale bias-controlled dataset \cite{barbu2019objectnet}. The Objectnet dataset contains 50,000 test images with a total of 313 classes, of which 113 overlap with Imagenet. Uncertainty calibration under domain drift was evaluated for CIFAR-10 based on affine transformations, and for Imagenet based on the perturbations introduced by \cite{hendrycks2019benchmarking} as well as the overlapping classes in Objectnet.\\
In addition, we quantify the quality of predictive uncertainty for truly OOD scenarios by computing the predictive entropy and distribution of confidence scores. We use complete OOD datasets as well as data perturbed at the highest level. In these scenarios we expect entropy to reach maximum levels, since the model should transparently communicate it "does not know" via low and unbiased confidence scores.

\subsection{Improving calibration under domain drift}
Existing methods for post-hoc uncertainty calibration are based on a validation set, which is drawn from the same generative distribution $\pi(X,Y)$ as the training set and the test set. Using these data to optimize a post-hoc calibration method results in low calibration errors for data drawn from $\pi(X,Y)$. If we would like to generalise this approach to calibration under domain drift, we need access to samples from the generative distribution along the axis of domain drift. However, such robustness under domain drift is a challenging requirement since in practice for a $D$-dimensional input domain drift can occur in any of the $2^D$ directions in $\{-1, 1\}^D$, and to any degree. Manifestations of such domain shifts include for example changes in viewpoint (Fig. 1), lighting condition, object rotation or background.\\
To obtain a transformed validation set representing a generic domain shift, we therefore sample domain drift scenarios by randomly choosing direction and magnitude of the domain drift. We use these scenarios to perturb the validation set and, taken together, simulate a generic domain drift. More specifically, we first choose a random direction $d_t \in \{-1, 1\}^D$. Next, we sample from a set of 10 noise levels $\epsilon$ covering the entire spectrum from in-domain to truly out-of domain. Each noise level corresponds to the variance of a Gaussian which in turn is used to sample the magnitude of domain drift. Since level and direction of domain shift are not known a priori, we argue that an image transformation using such Gaussian noise results in a generic validation set in the spirit of the central limit theorem: we emulate complex domain shifts in the real world by performing additive random image transformations, which in turn can be approximated by a Gaussian distribution.\\
We optimise $\epsilon$ in a dataset-specific manner such that the accuracy of the pre-trained model decreases linearly in 10 steps to random levels (See Appendix for detailed algorithm).\\
\begin{table*}
\begin{tabular}{lrrrrrrrrrrr}
\toprule
{} &      Base &        TS &       ETS &    TS-IR    &        IR &     IRM &      TS-P &     ETS-P &   TS-IR-P   &      IR-P &   IRM-P \\

\midrule
CIFAR VGG19             &  0.323 & 0.158 &  0.152 &  0.173 &  0.176 &  0.167 &  0.053 &  0.057 &  0.051 &  0.049 &  \textbf{0.044} \\
CIFAR ResNet50          &  0.202 &  0.176 &  0.171 &  0.191 &  0.190 &  0.179 &  0.083 &  0.090 &  0.092 &  0.093 &  \textbf{0.076} \\
CIFAR DenseNet121       &  0.206 &  0.151 &  0.145 &  0.166 &  0.168 &  0.152 &  0.135 &  0.122 &  0.103 &  \textbf{0.088} &  0.120 \\
CIFAR MobileNetv2       &  0.159 &  0.150 &  0.141 &  0.165 &  0.165 &  0.147 &  0.107 &  0.125 &  0.094 &  \textbf{0.079} &  0.108 \\
\midrule
ImgNet ResNet50       &  0.130 &  0.049 &  0.064 &  0.134 &  0.142 &  0.072 &  0.050 &  0.041 &  \textbf{0.033} &  0.037 &  0.041 \\
ImgNet ResNet152      &  0.129 &  0.043 &  0.049 &  0.127 &  0.135 &  0.062 &  0.037 &  0.034 &  \textbf{0.028} &  0.039 &  0.045 \\
ImgNet VGG19          &  0.057 &  0.045 &  0.047 &  0.120 &  0.122 &  0.051 &  0.093 &  0.075 &  0.064 &  \textbf{0.029} &  0.047 \\
ImgNet Den.Net169    &  0.117 &  0.044 &  0.040 &  0.127 &  0.133 &  0.057 &  0.024 &  \textbf{0.023} &  0.026 &  0.045 &  0.050 \\
ImgNet Eff.NetB7 &  0.092 &  0.135 &  0.085 &  0.131 &  0.132 &  0.074 &  0.074 &  0.047 &  \textbf{0.038} &  0.049 &  0.058 \\
ImgNet Xception       &  0.205 &  0.068 &  0.042 &  0.109 &  0.130 &  0.076 &  0.060 &  0.031 &  \textbf{0.031} &  0.101 &  0.101 \\
ImgNet Mob.Netv2    &  0.063 &  0.143 &  0.114 &  0.186 &  0.181 &  0.107 &  0.099 &  0.074 &  0.066 &  \textbf{0.046} &  0.069 \\
\bottomrule
\end{tabular}
\caption{Mean expected calibration error across all test domain drift scenarios (affine transformations for CIFAR-10 and perturbations proposed in \cite{hendrycks2019benchmarking} for Imagenet). For all architectures our approach of using a perturbed validation set outperformed baseline  post-hoc calibrators}\label{tab:mean_ece}
\end{table*}
\begin{figure}[t!]
	\centering
	\begin{subfigure}[t]{0.25\textwidth}
		\centering
    	\includegraphics[width=\textwidth]{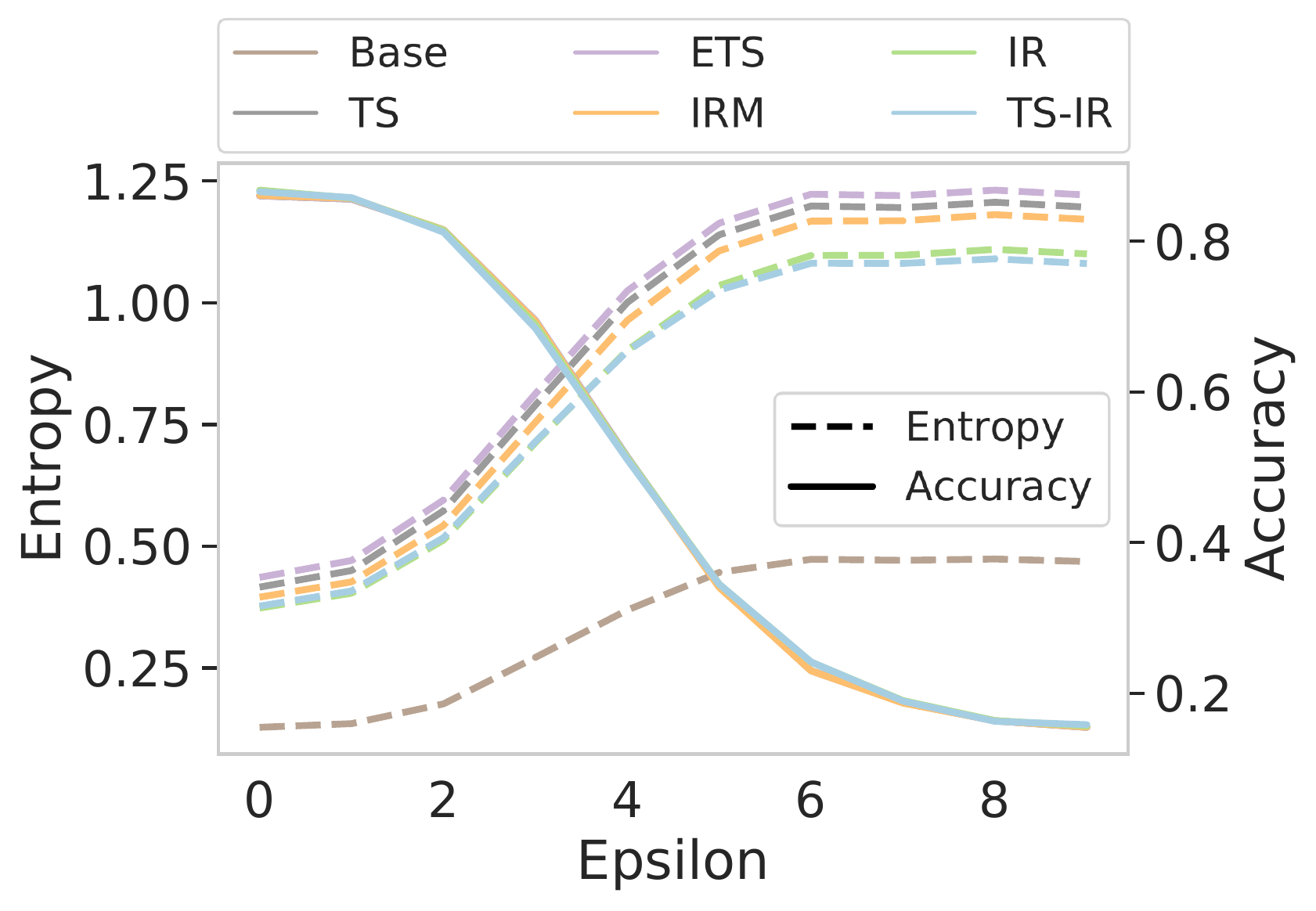}
		\caption{Accuracy and entropy for CIFAR-10 data}
	\end{subfigure}
	\begin{subfigure}[t]{0.214\textwidth}
		\centering
		\includegraphics[width=\textwidth]{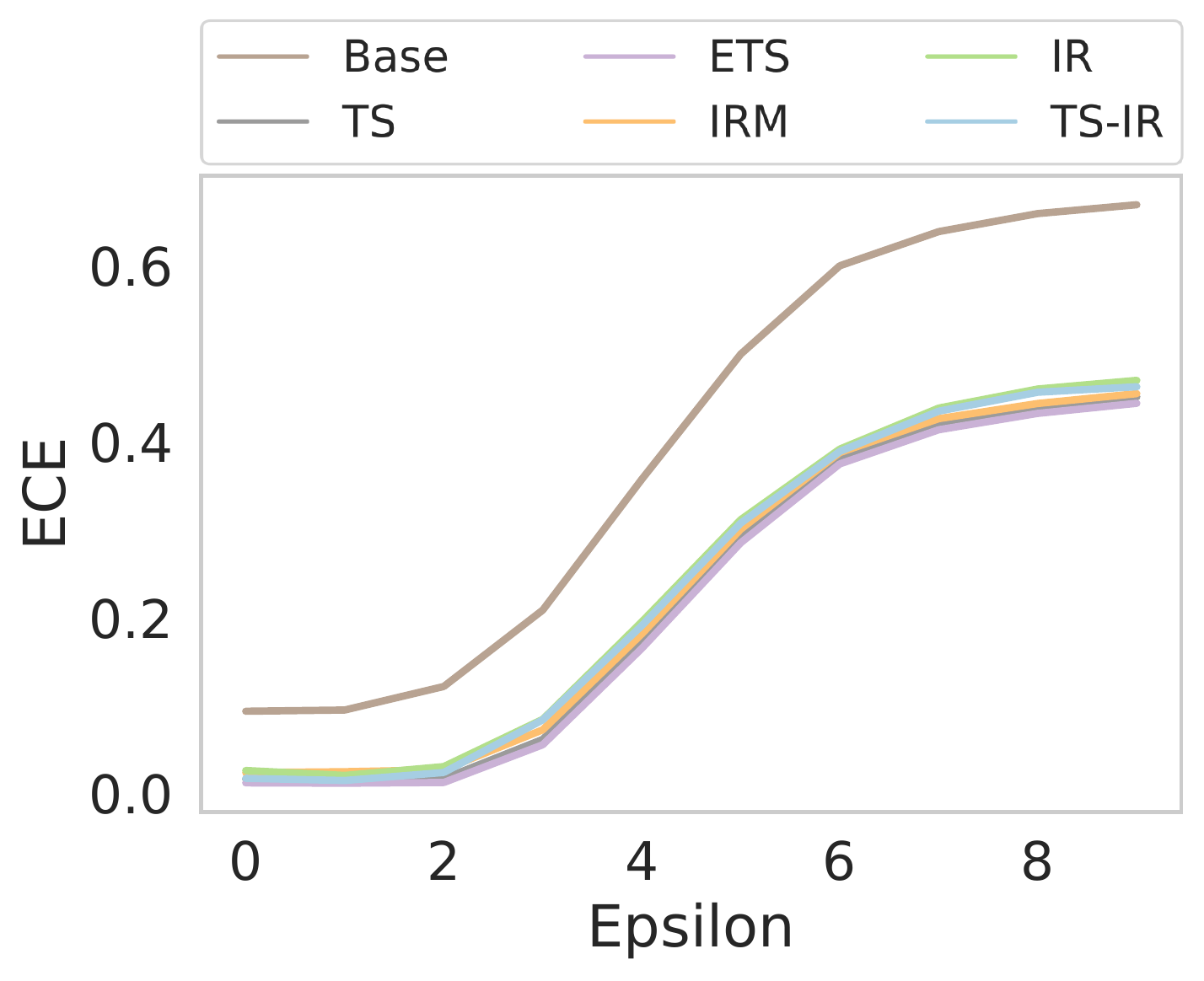}
		\caption{ECE for CIFAR-10 data}
	\end{subfigure}

	\caption{Model performance in terms of accuracy, entropy and expected calibration error for CIFAR-10 data for perturbation shear. (a) As expected accuracy degrades with increasing perturbation to almost random levels for all models. (b) While entropy increases with increasing perturbation strength Epsilon for all models, ECE also increases for all models, indicating a mis-match between confidence and accuracy.}
\label{fig:line_mnist}
\end{figure}

\begin{figure}
	\centering
	\begin{subfigure}[t]{0.23\textwidth}
		\centering
    	\includegraphics[width=\textwidth]{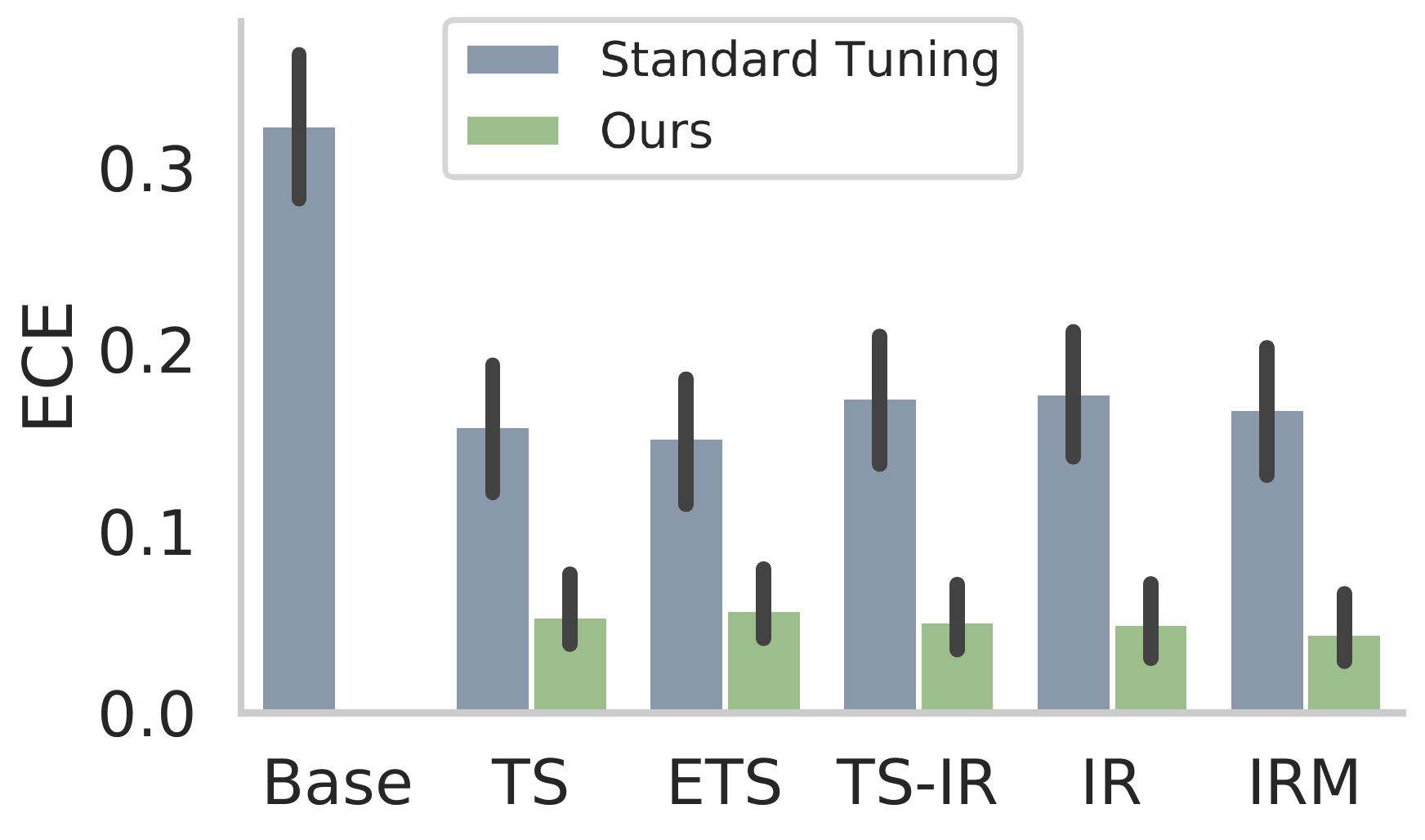}
		\caption{CIFAR-10}
	\end{subfigure}
	\begin{subfigure}[t]{0.23\textwidth}
		\centering
		\includegraphics[width=\textwidth]{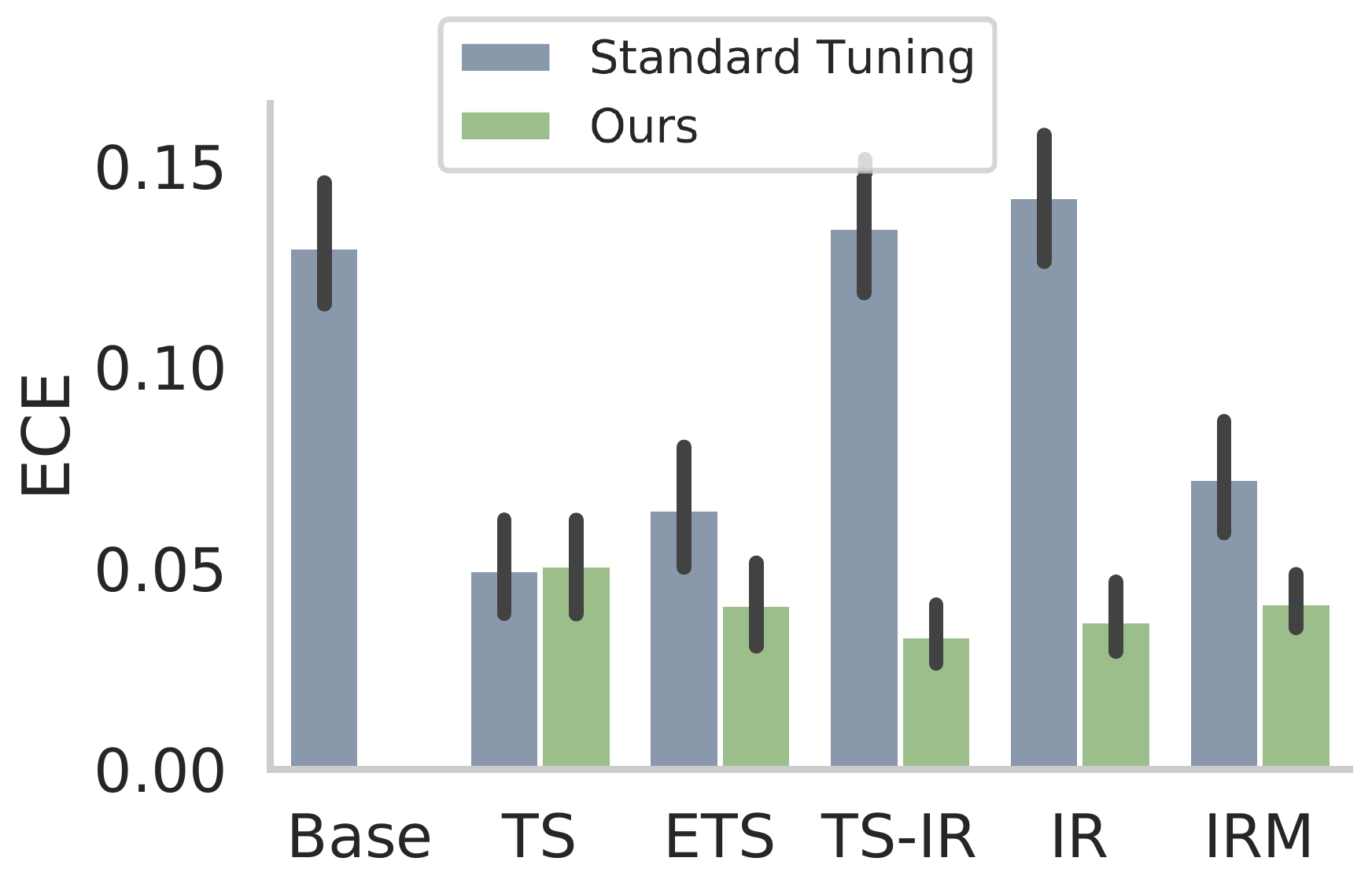}
		\caption{Imagenet}
	\end{subfigure}
	\caption{Mean expected calibration error, averaged over all test scenarios and levels of domain drift. Using our proposed calibration approach improves the overall calibration error for all post-hoc calibrators.}
\label{fig:meanece_cifar_imagenet}
\end{figure}

In summary, we sample a domain shift scenario using Gaussian noise for each sample in the validation set, thereby generating a perturbed validation set. We then tune a given post-hoc uncertainty calibration method based on this perturbed validation set and obtain confidence scores that are calibrated under domain drift. All in all, we simulate domain drift scenarios and use the resulting perturbed validation set to tune existing post-hoc uncertainty calibration methods. We hypothesize that this facilitates calibrated predictions of neural networks under domain drift.\\
We refer to tuning a post-hoc calibrator using the perturbed validation set by a suffix "-P", e.g. IR-P stands for Isotonic Regression tuned on the perturbed validation set.

\section{Experiments and results}

\begin{figure*}
	\centering
	\begin{subfigure}[t]{0.9\textwidth}
		\centering
    	\includegraphics[width=\textwidth]{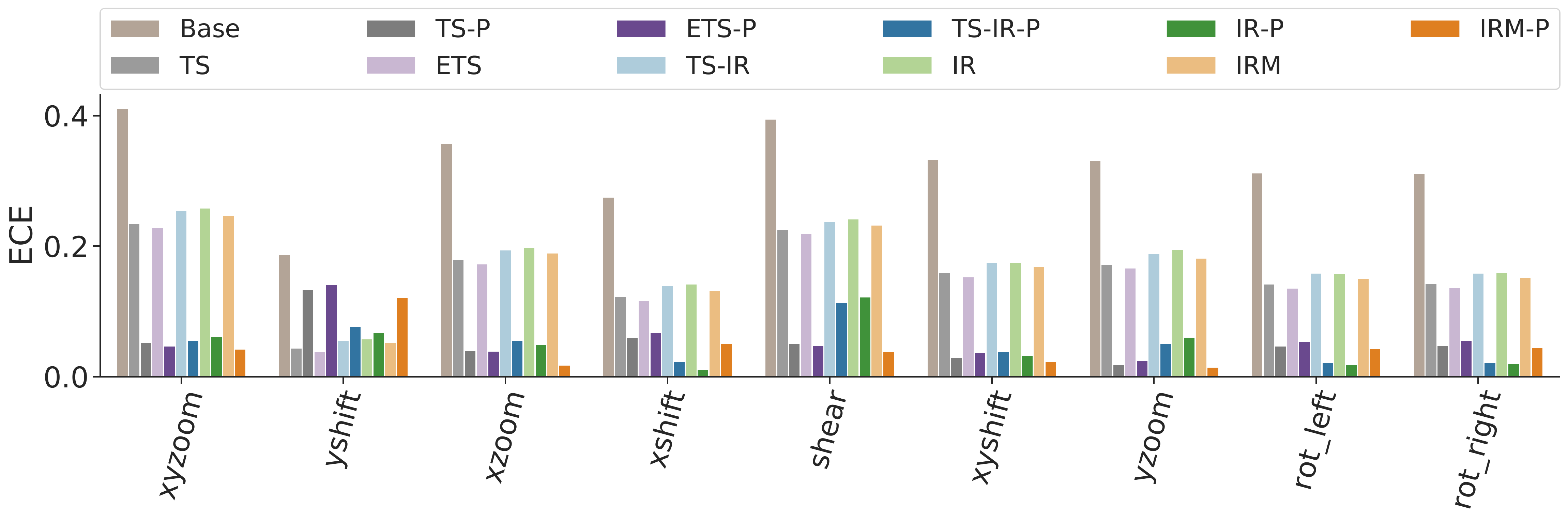}
		\caption{CIFAR-10}
	\end{subfigure}
	\begin{subfigure}[t]{0.9\textwidth}
		\centering
		\includegraphics[width=\textwidth]{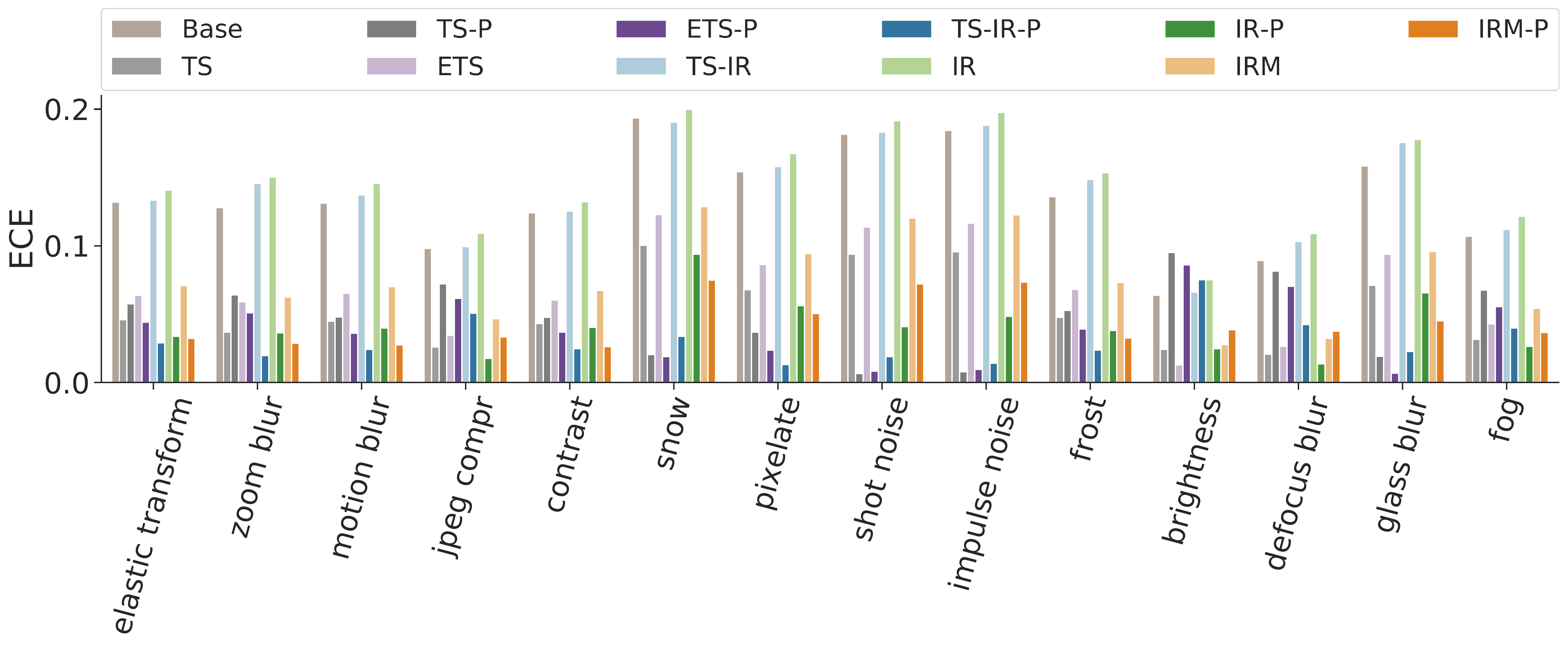}
		\caption{Imagenet}
	\end{subfigure}
	\caption{Micro-averaged calibration error for individual test domain shift scenarios}
\label{fig:ece_per}
\end{figure*}

We first illustrate limitations of post-hoc uncertainty calibration methods in domain drift scenarios using CIFAR-10. We show that while excellent in-domain calibration can be achieved using standard baselines, the quality of uncertainty decreases with increasing domain shift for all methods, resulting in highly overconfident predictions for images far away from the training domain.\\
Next, we show on a variety of architectures and datasets that replacing the validation set by a transformed validation set as outlined in section 4.2, substantially improves calibration under domain shift. We further assess the effect of our new approach on in-domain calibration and demonstrate that for selected post-hoc calibration methods, in-domain calibration can be maintained at competitive levels. Finally, we show that our tuning approach results in better uncertainty awareness in truly OOD settings.
\subsection{Post-hoc calibration results in overconfident predictions under domain drift}
We tuned all post-hoc calibration baseline methods on the CIFAR-10 validation set and first assessed in-domain calibration on the test set. As expected, calibration improves for all baselines over the uncalibrated network predictions in this in-domain setting. Next, we assessed calibration under domain drift by generating a perturbed test where we apply different perturbations (e.g. rotation, shear, zoom) to the images. We increased perturbation strength (i.e. shear) in 9 steps until reaching random accuracy. Figure \ref{fig:line_mnist} illustrates, that while entropy increases with intensifying shear for all models, ECE also increases for the entire set of models. This reveals a mis-match between confidence and accuracy that increases with increasing domain drift. 
\begin{figure}[ht]
	\centering
	\includegraphics[width=0.37\textwidth]{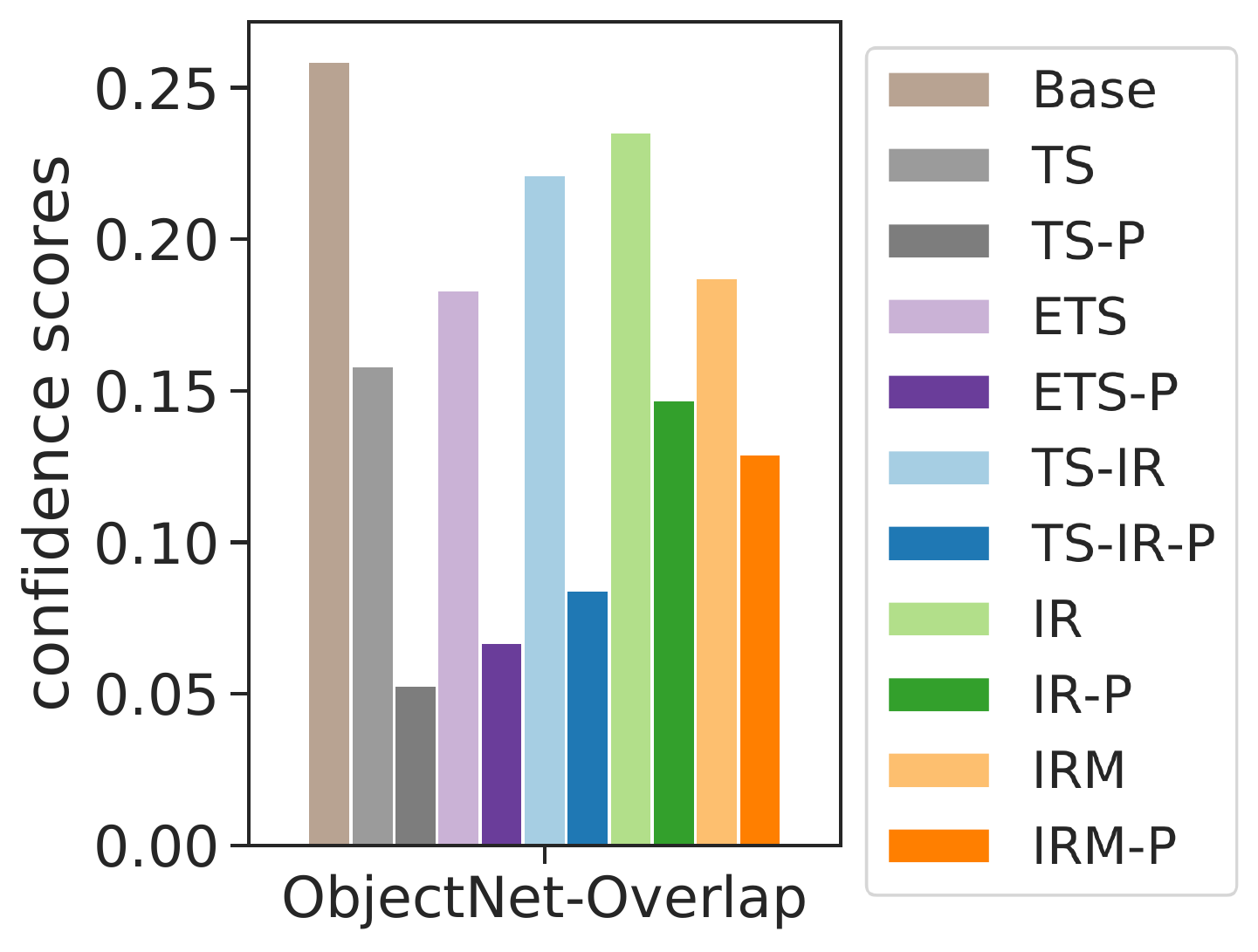} 
	\caption{ECE for all overlapping classes contained in Objectnet dataset. Our tuning approach improves calibration for all post-hoc calibrators (Resnet50 trained on Imagenet dataset).}
	\label{fig:ECE_obj}
\end{figure}

\begin{figure*}[t!]
	\centering
	\begin{subfigure}[t]{0.48\textwidth}
		\centering
    	\includegraphics[width=\textwidth]{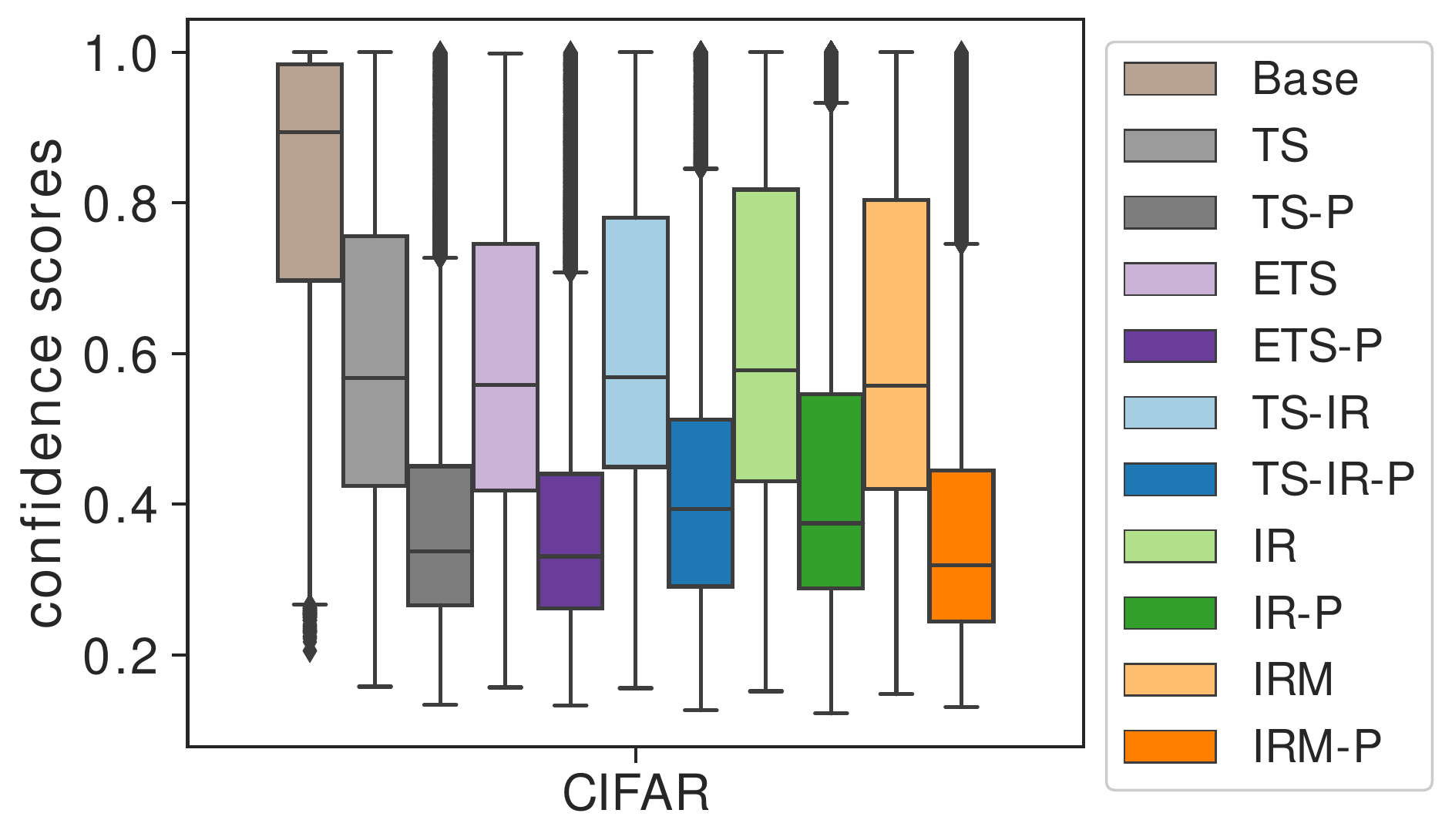}
		\caption{OOD predictions CIFAR}
	\end{subfigure}
	\begin{subfigure}[t]{0.48\textwidth}
		\centering
		\includegraphics[width=\textwidth]{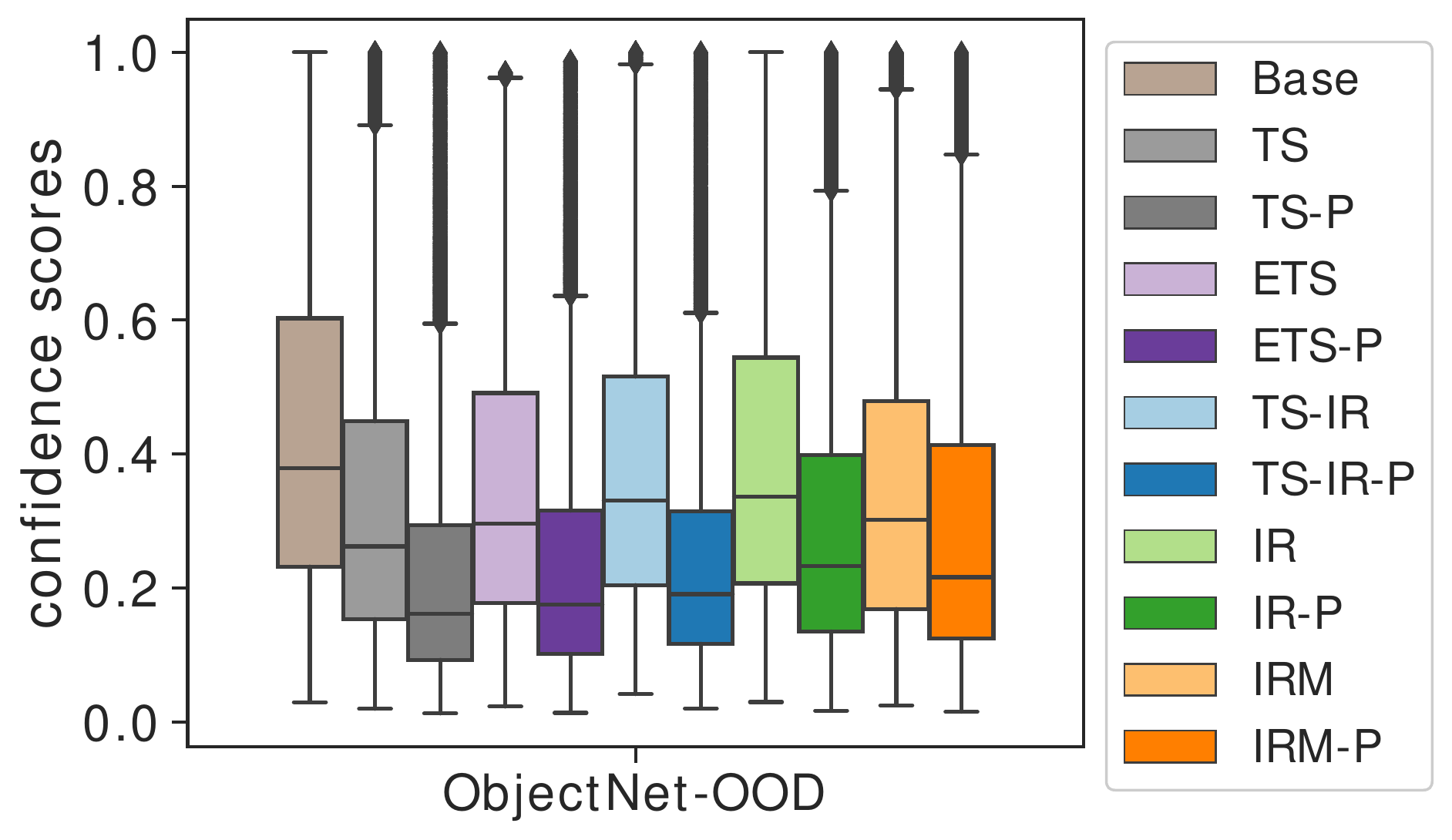}
		\caption{OOD predictions Imagenet (Resnet50).}
	\end{subfigure}

	\caption{Distribution of confidence scores for out-of-domain predictions. (a) Confidence scores for our tuning strategy (-P) are substantially lower for the highest perturbation level compared to standard tuning, reflecting that our approach yields well calibrated predictions also for truly OOD samples (b) Confidence scores for OOD dataset Objecnet (non-overlapping classes) reveals that our approach results in substantially more uncertainty aware predictions.}
\label{fig:ood}
\end{figure*}
\begin{figure*}
	\centering
	\begin{subfigure}[t]{0.85\textwidth}
		\centering
    	\includegraphics[width=\textwidth]{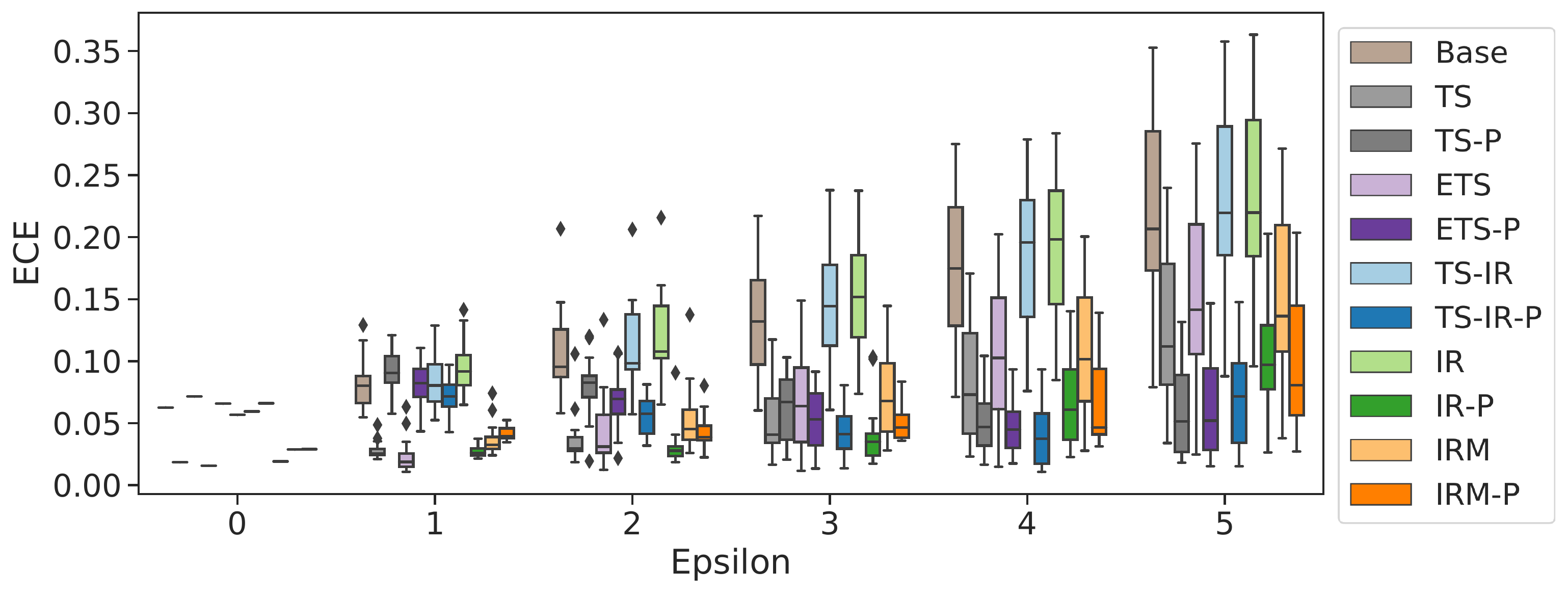}
		\caption{ECE Imagenet}
	\end{subfigure}
	\begin{subfigure}[t]{0.85\textwidth}
		\centering
		\includegraphics[width=\textwidth]{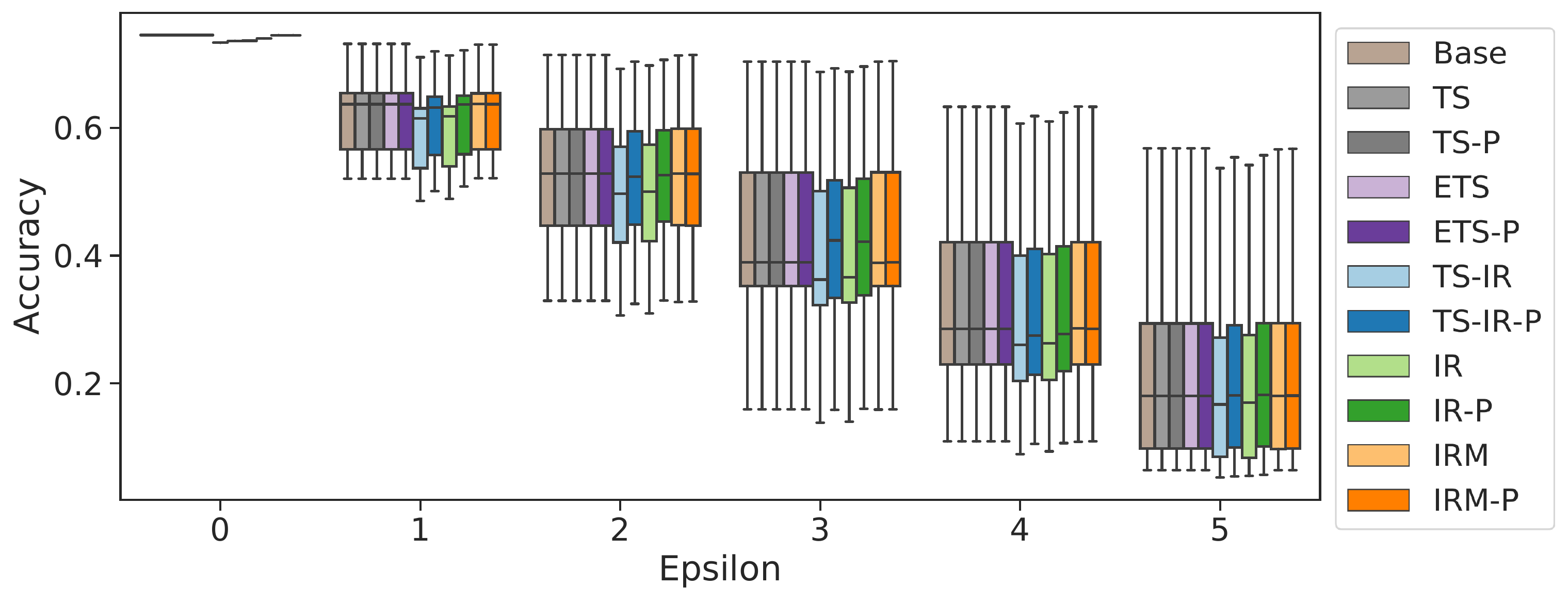}
		\caption{Accuracy Imagenet}
	\end{subfigure}
	\caption{Dependency of accuracy and and ECE on level of domain shift. Accuracy decreases for all methods with increasing levels of domain shift. For standard post-hoc calibrators ECE also increases, but our approach of tuning on a perturbed validation set results in good calibration throughout all levels, especially for IR-P.}
\label{fig:ece_eps}
\end{figure*}
\subsection{Perturbed validation sets improve calibration under domain drift}
Next, we systematically assessed whether calibration under domain drift can be improved by tuning post-hoc calibration methods on a transformed validation set. 
To this end, we calibrate various neural network architectures on CIFAR-10 and Imagenet. For CIFAR-10, we first train VGG19 \cite{simonyan2014very}, ResNet50 \cite{he2016deep}, DenseNet121 \cite{huang2017densely} and MobileNetv2 \cite{sandler2018mobilenetv2} models. For Imagenet we used 7 pre-trained models provided as part of tensorflow, namely ResNet50, ResNet152, VGG19, DenseNet169, EfficientNetB7 \cite{tan2019efficientnet}, Xception \cite{chollet2017xception} and MobileNetv2.\\
For all neural networks we then tuned 2 sets of post-hoc calibrators: one set was tuned in a standard manner based on the validation set, the second set was tuned with the proposed method based on the perturbed validation set. We then evaluate both sets of calibrators under various domain drift scenarios that were not seen during training, as well as in terms of in-domain calibration.\\
We observed that for all post-hoc calibrators tuning on a perturbed validation set resulted in an overall lower calibration error when testing across all domain drift scenarios, Table 1, Fig. \ref{fig:meanece_cifar_imagenet}).
Figure \ref{fig:ece_per} illustrates for VGG19 trained on CIFAR-10 and Resnet50 trained on Imagenet that this improvement was consistent for individual domain drift scenarios not seen during training.

\paragraph{Real-world domain shift} We next assessed the effect of our proposed tuning strategy on a real-world domain-shift. To this end, we computed ECE on the Objectnet test dataset. This comprises a set of images showing objects also present in Imagenet with different viewpoints, on new backgrounds and different rotation angles. As for artificial image perturbations, we found that our tuning strategy resulted in better calibration under domain drift compared to standard tuning, for all post-hoc calibration algorithms (Fig. \ref{fig:ECE_obj}).

\paragraph{Dependency on magnitude of domain drift}
We next focused on the Imagenet to assess how calibration depends on the amount of domain shift. We observe that while standard post-hoc calibrators yield very low in-domain calibration errors ((Fig. \ref{fig:ece_eps}, ECE at epsilon 0), predictions become increasingly overconfident with increasing domain shift. In contrast, all methods tuned using our approach had a low calibration error for large domain shifts. Notably, we observed two types of behaviour for in-domain calibration and small domain shift. One set of methods - TS-P, ETS-P and TS-IR-P - had a substantially increased calibration error for in-domain settings compared to standard calibration methods TS, ETS and TS-IR respectively. For these methods ECE at low levels of domain shift was comparable to that of uncalibrated models, and \it decreased \rm substantially with domain shift (in contrast to uncalibrated models or standard calibrators where ECE \it increases \rm with domain shift). However, IR-P did not show a worse in-domain calibration compared to standard post-hoc calibrators when tuned using our approach. Notably, it yielded a calibration error comparable to state-of-the-art calibrators for in-domain settings, while substantially improving calibration for increasing levels of domain shift. We further observed that IRM-P, an accuracy preserving version of IR-P, had a small but consistently worse in-domain classification error than IR-P, but performed substantially better than TS-based methods for small domain shifts.\\
One key difference between IR and TS-based methods, is that the latter methods are accuracy preserving, while IR-P (and IR) can potentially change the ranking of predictions and thus accuracy. However, for Imagenet we only observed minor changes in accuracy for IR-based methods, irrespective of tuning strategy (Fig. \ref{fig:ece_per} (b), Fig. \ref{fig:ece_eps} (b)). We hypothesize that the systematic difference in calibration behaviour is due to the limitations of expressive power of TS-based accuracy-preserving methods: first, by definition the ranking between classes in a multi-class prediction does not change after applying this family of calibrators and second only one parameter (for TS) or 4 parameters (ETS) have to capture the potentially complex behaviour of the calibrator. While this has important advantages for practitioners, it can also result in a lack of expressive power which may be needed to achieve calibration for all levels of domain drift, ranging from in-domain to truly out-of-domain. Finally, we observed that a variance reduction via temperature scaling \cite{kumar2019verified} before IR (TS-IR) was also not beneficial for calibration under small levels of domain shift. 

\subsection{Out-of-distribution scenarios}
To further investigate the behaviour of post-hoc calibrators for truly OOD scenarios, we analysed the distribution of confidence scores for a completely OOD dataset for Imagenet, using the 200 non-overlapping classes in Objectnet. This revealed that all post-hoc calibrators resulted in significantly less overconfident predictions when tuned using our approach. We observed a similar behaviour for CIFAR-10 with VGG19, when assessing the distribution of confidence scores at the highest perturbation level across all perturbations (Fig. \ref{fig:ood}).

\section{Discussion and conclusion}
We present a simple and versatile approach for tuning post-hoc uncertainty calibration methods. We demonstrate that our new approach, when used in conjunction with isotonic regression-based methods (IR or IRM), yields well-calibrated predictions in the case of any level of domain drift, from in-domain to truly out-of-domain scenarios. Notably, IR-P and IRM-P maintain their calibration performance for in-domain scenarios compared to standard isotonic regression (IR and IRM). In other words, our experiments suggest that when using our IR(M)-P approach, there is only a minor trade-off between choosing a model that is either well calibrated in-domain vs. out-of-domain. In contrast, methods based on temperature scaling may not have enough expressive power to achieve good calibration across this range: standard tuning results in highly overconfident OOD predictions and perturbation-based tuning results in calibration errors comparable to uncalibrated models for in-domain predictions. However, when averaging across all domain drift scenarios, overall calibration for TS-P and ETS-P still improves substantially over standard TS and ETS. Consequently, for use-cases requiring an accuracy preserving method and reliable uncertainty estimates especially for larger levels of domain shift, TS-P and ETS-P are good options.\\
We further observe this trade-off between accuracy preserving properties and calibration error for IR-based methods. While IRM-P has accuracy-preserving properties, overall calibration errors are higher than for IR-P in particular for small domain shifts.\\

Our perturbation-based tuning can be readily applied to any post-hoc calibration method. When used in combination with an expressive non-parametric method such as IR, this results in well calibrated predictions not only for in-domain and small domain shifts, but also for truly OOD samples. This is in stark contrast to existing methods with standard tuning, where performance in terms of calibration and uncertainty-awareness degrades with increasing domain drift.

\section{Acknowledgements}
This work was supported by the Munich Center for Machine Learning and has been funded by the German Federal Ministry of Education
and Research (BMBF) under Grant No. 01IS18036B.

{\small
\bibliographystyle{ieee_fullname}
\bibliography{references}
}

\clearpage

\appendix








\onecolumn




\section{Appendix Summary}
We provide further details on the implementation of our algorithm as well as additional results. This appendix is structured as follows. 
\begin{itemize}
    \item In section B, we first formalize our algorithm in Algorithm 1. We then provide more details on the test perturbations used for our analyses, along with their parameter sets. 
    \item In section C, we report supplementary results, including additional metrics as well as data on additional baselines and additional experiments on the robustness of our findings.
    \item In section D, we consolidate our results into a brief recommendation for practitioners.
\end{itemize}

\section{Additional implementation details}
\subsection{Algorithm}
    \begin{algorithm}[H]
    	\caption{Tuning of a post-calibration method for domain shift scenarios\\ \textbf{Input}: Classification model $Y=f(X)$, validation set $(X, Y)$, number of perturbation levels $N$, number of classes $C$, initial parameter $\varepsilon_{\mathrm{init}}$.}
    	\begin{algorithmic}[1]
            \STATE Compute min and max accuracy: $acc_{\min}=1/C$; $acc_{\max}=\mathrm{acc}(f(X),Y)$
            \STATE Compute $N$ evenly spaced accuracy levels $A=\{acc_{\min},acc_{\min}+\frac{acc_{\max}-acc_{\min}}{N-1},\dots, acc_{\max}\}$ 
            \STATE Initialise empty perturbed validation set $(X_\mathcal{E},Y_\mathcal{E})$
            \FOR{$i$ in 1:N}
            \IF{$i=1$}
            \STATE Set $\varepsilon_{i}=\varepsilon_{\mathrm{init}}$
            \ENDIF
            \STATE Compute $X_{\varepsilon_i}=X+\mathcal{N}(0,\varepsilon_i)$, by drawing a sample from a Gaussian $\mathcal{N}(0,\varepsilon_i)$ with variance $\varepsilon_i$ for every pixel $(j,k)$ in every image $x_{j,k} \in X$
            \STATE Minimize $\mathrm{acc}(f(X_{\varepsilon_i}),Y) - A_{i}$ with respect to $\varepsilon_{i}$,  using  a Nelder-Mead optimizer
            \STATE Compute $X_{\varepsilon_i}=X+\mathcal{N}(0,\varepsilon_i)$ using optimized $\varepsilon_i$
            \STATE Add $(X_{\varepsilon_i},Y)$ to $(X_\mathcal{E},Y_\mathcal{E})$
            \IF{$i<N$}
            \STATE Initialise $\varepsilon_{i+1}=\varepsilon_{i}$
            \ENDIF
            \ENDFOR
        	\STATE Tune post-processing method on $(X_\mathcal{E},Y_\mathcal{E})$
        \end{algorithmic}
    \end{algorithm}

\subsection{Perturbation strategies}

For the affine test  perturbation strategies (Table \ref{tab:pert_para1}) we chose 10 levels of perturbation with increasing perturbation strength until random levels of accuracy were reached (or parameters could not be increased any further). We started all test perturbation sequences at no perturbation and list specific levels of perturbation in Table \ref{tab:pert_para1}.\\
For Imagenet corruptions, we follow \cite{hendrycks2019benchmarking} and report test accuracy as well as accuracy under maximum domain shift in Table \ref{tab:pert_para}.

\begin{table}[b]
\centering
\caption{For rotation, perturbation is the (left or right) rotation angle in degrees, shift is measured in pixels in x or y direction, for shear the perturbation is measured as shear angle in counter-clockwise direction in degrees, for zoom the perturbation is zoom in x or y direction. }

\begin{tabular}{l l l l l l l l l l l}
\toprule
 Perurbation &  \multicolumn{10}{c}{Perturbation-specific parameter}\\
 \midrule
rot left  & 0 & 350 & 340 & 330 & 320 & 310 & 300 & 290 & 280 & 270 \\ 
rot right  & 0 & 10 & 20 & 30 & 40 & 50 & 60 & 70 & 80 & 90 \\ 
shear  & 0 & 10 & 20 & 30 & 40 & 50 & 60 & 70 & 80 & 90 \\ 
xyshift  & 0 & 2 & 4 & 6 & 8 & 10 & 12 & 14 & 16 & 18 \\ 
xshift & 0 & 2 & 4 & 6 & 8 & 10 & 12 & 14 & 16 & 18 \\ 
xyshift  & 0 & 2 & 4 & 6 & 8 & 10 & 12 & 14 & 16 & 18 \\ 
xyzoom  & 1 & 0.90 & 0.80 & 0.70 & 0.60 & 0.50 & 0.40 & 0.30 & 0.20 & 0.10 \\ 
xzoom & 1 & 0.90 & 0.80 & 0.70 & 0.60 & 0.50 & 0.40 & 0.30 & 0.20 & 0.10 \\ 
yzoom & 1 & 0.90 & 0.80 & 0.70 & 0.60 & 0.50 & 0.40 & 0.30 & 0.20 & 0.10 \\ 
\end{tabular}
\label{tab:pert_para1}
\end{table}

\begin{table}
\centering
\caption{Accuracies for Imagenet perturbations in-domain and with maximum shift.}

\begin{tabular}{l l l l l l l l l l l}
\toprule
 Perurbation &  \multicolumn{2}{c}{Accuracy}\\
 & In-Domain & Max Domain-Shift  \\ 
 \midrule
shot noise & 0.7452 & 0.07752  \\ 
impulse noise & 0.7452 & 0.07104  \\ 
defocus blur & 0.7452 & 0.14784  \\ 
glass blur & 0.7452 & 0.06904  \\ 
motion blur & 0.7452 & 0.09696  \\ 
zoom blur & 0.7452 & 0.22864  \\ 
snow & 0.7452 & 0.17776  \\ 
frost & 0.7452 & 0.25016  \\ 
fog & 0.7452 & 0.40912  \\ 
brightness & 0.7452 & 0.56776  \\ 
contrast & 0.7452 & 0.06416  \\ 
elastic transform & 0.7452 & 0.14480  \\ 
pixelate & 0.7452 & 0.19216  \\ 
jpeg compression & 0.7452 & 0.41136  \\ 
gaussian blur& 0.7452 & 0.10016  \\ 
saturate & 0.7452 & 0.47952  \\ 
spatter & 0.7452 & 0.30808  \\ 
speckle noise & 0.7452 & 0.18296  \\ 
\end{tabular}
\label{tab:pert_para}
\end{table}

\newpage
\section{Additional results}
\subsection{Additional baselines}
In addition to the state-of-the-art post-calibrators analysed in detail in the main paper, we also assessed the effect of tuning based on a perturbed validation set for additional baselines. Here, we report results for CIFAR-10 for Platt scaling \cite{Platt99probabilisticoutputs}, histogram binning \cite{zadrozny2001obtaining} and a recently proposed approach combining Platt scaling with histogram binning (PBMC) \cite{kumar2019verified}.\\
Table \ref{tab:addbase} reveals that also these baselines benefit from tuning on a perturbed validation set; note however that overall ECE was consistently higher for these baselines compared to IR-P, for all architectures.

\begin{table}[h]
    \caption{Mean micro-average ECE across all affine test perturbations for the additional baselines.}
    \centering
\begin{tabular}{lrrrrrrr}
\toprule
{} &      Base &        PS &        HB &      PBMC &      PS-P &      HB-P &    PBMC-P \\
\midrule
CIFAR VGG19       &  0.323 &  0.173 &  0.254 &  0.211 &  \textbf{0.075} &  0.086 &  0.101 \\
CIFAR ResNet50    &  0.202 &  0.211 &  0.220 &  0.210 &  0.181 &  0.101 &  \textbf{0.099} \\
CIFAR Den.Net121 &  0.206 &  0.177 &  0.205 &  0.191 &  0.109 &  \textbf{0.096} &  0.105 \\
CIFAR Mob.NetV2 &  0.159 &  0.180 &  0.191 &  0.187 &  0.182 &  0.099 &  \textbf{0.098} \\
\bottomrule
\end{tabular}
    \label{tab:addbase}
\end{table}

\subsection{Additional metrics}
In addition to the expected calibration error as reported in the main paper, we also compute a debiased ECE, recently proposed in \cite{kumar2019verified}, that can be more robust than the standard definition of ECE. Also with this measure, our approach improves all baselines consistently, with IRM-P, IR-P and TS-IR-P performing best (Table \ref{tab:deECE}).

\begin{table}[h]
    \caption{Debiased ECE for all baselines for CIFAR-10 and Imagenet.}

    \centering
\begin{tabular}{lrrrrrrrrrrr}
\toprule
{} &      Base &        TS-P &     ETS-P &  TS-IR-P  &      IR-P &     IRM-P\\
\midrule
CIFAR VGG19             &  0.371 &  0.065 &  0.070 &  0.061 &  0.058 &  \textbf{0.054} \\
CIFAR ResNet50          &  0.221 &  0.099 &  0.110 &  0.101 &  0.101 &  \textbf{0.089} \\
CIFAR DenseNet121       &  0.230 &  0.162 &  0.148 &  0.118 &  \textbf{0.100} &  0.141 \\
CIFAR MobileNetv2       &  0.176 &  0.129 &  0.152 &  0.109 &  \textbf{0.089} &  0.132 \\
\midrule
ImgNet ResNet50       &  0.144 &  0.058 &  0.047 &  \textbf{0.042} &  \textbf{0.042} &  0.050 \\
ImgNet ResNet152      &  0.144 &  0.042 &  0.039 &  \textbf{0.034} &  0.045 &  0.055 \\
ImgNet VGG19          &  0.064 &  0.108 &  0.087 &  0.079 &  \textbf{0.034} &  0.055 \\
ImgNet Den.Net169    &  0.129 &  0.027 &  0.027 &  \textbf{0.030} &  0.049 &  0.060 \\
ImgNet Eff.NetB7 &  0.109 &  0.089 &  0.055 &  \textbf{0.042} &  0.056 &  0.068 \\
ImgNet Xception       &  0.235 &  0.072 &  0.038 &  \textbf{0.035} &  0.119 &  0.122 \\
ImgNet MobileNetv2    &  0.070 &  0.113 &  0.084 &  0.080 &  \textbf{0.053} &  0.074 \\
\bottomrule
\end{tabular}
    \label{tab:deECE}
\end{table}

\begin{figure}[h]
	\centering
	\includegraphics[width=0.85\textwidth]{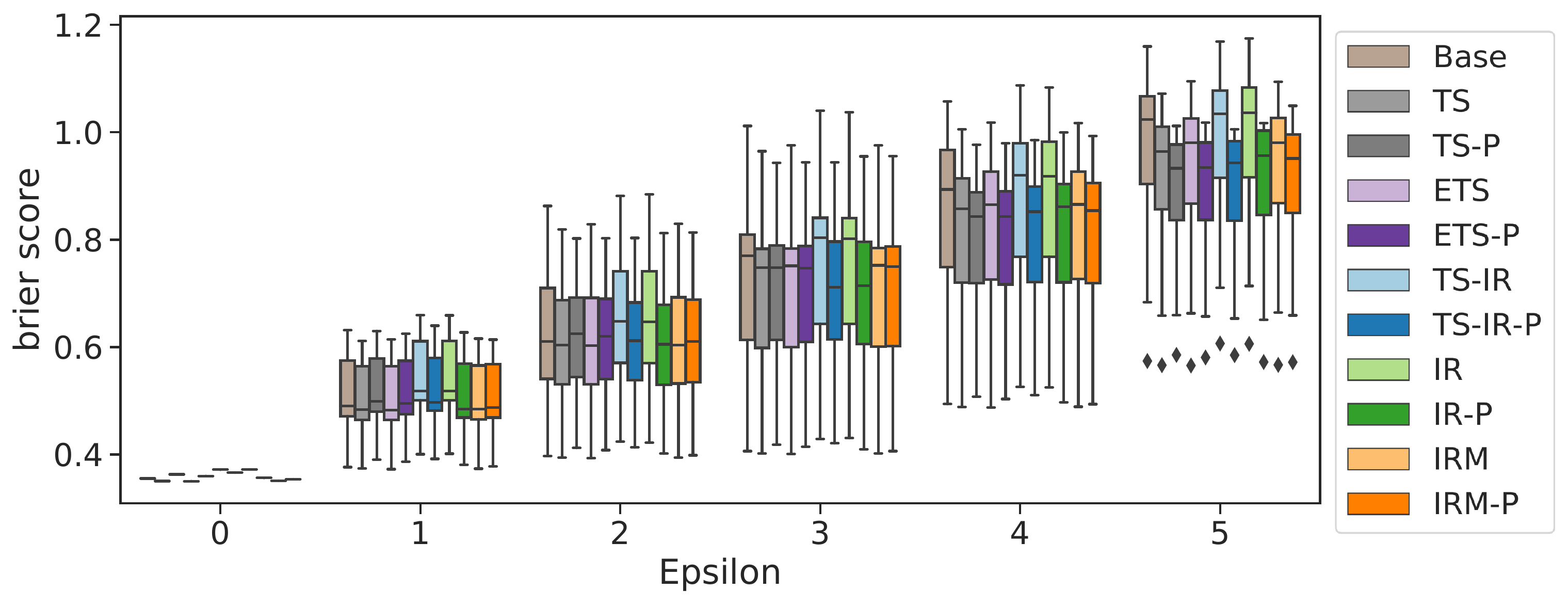}
	\caption{Brier score for Resnet50 trained on Imagenet}
	\label{fig:brier}
\end{figure}

\begin{figure}[h]
	\centering
	\includegraphics[width=0.85\textwidth]{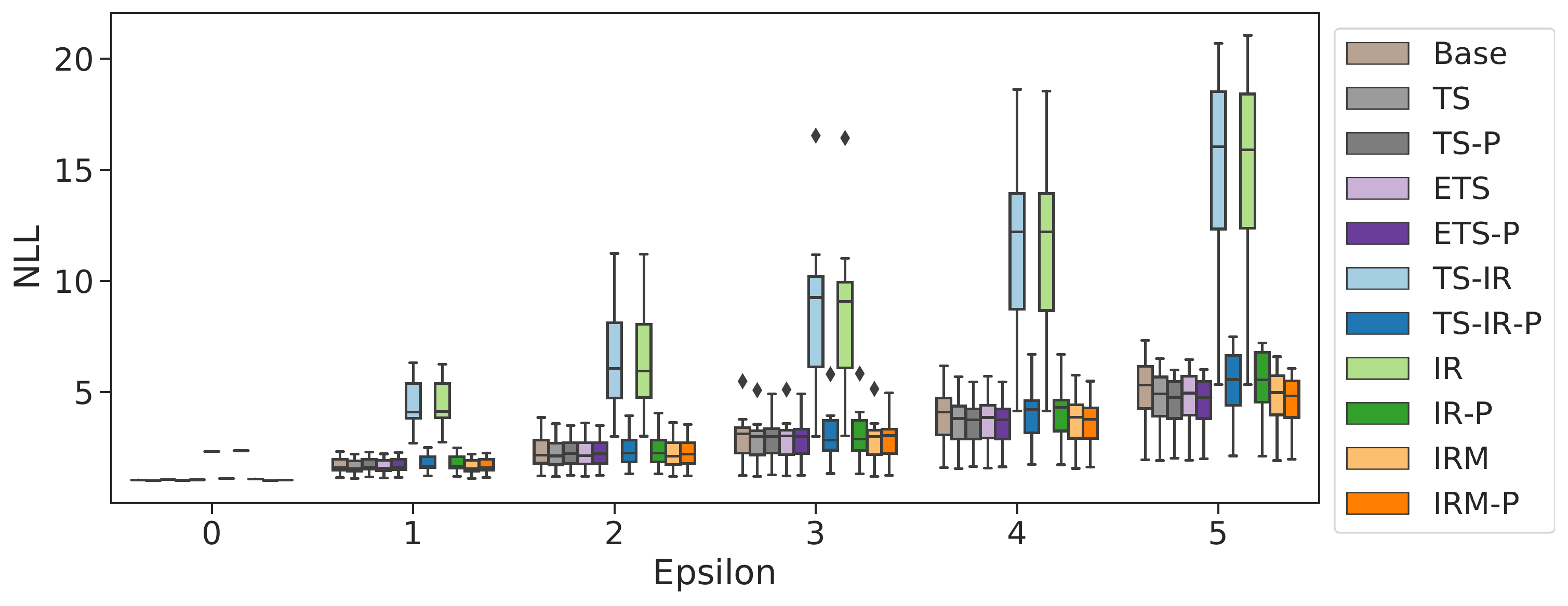}
	\caption{NLL for Resnet50 trained on Imagenet}
	\label{fig:nll}
\end{figure}

Furthermore, we also computed the negative log-likelihood as well as the Brier score for all post-calibrators. Again, our approach results in consistent improvements over the state-of-the-art also in terms of these metrics (Tables \ref{tab:nll} and \ref{tab:brier} and Figures \ref{fig:brier} and \ref{fig:nll}).\\
\begin{table}[h]
    \caption{NLL for all baselines for CIFAR-10 and Imagenet}

    \centering
\begin{tabular}{lrrrrrrrrrrr}
\toprule
{} &      Base &        TS &       ETS &     TS-IR  &         IR &    IRM&      TS-P &     ETS-P &   TS-IR-P &      IR-P &  IRM-P\\
\midrule
C-VGG19             &  2.49 &  1.49 &  1.47 &   1.86 &   1.88 &  1.55 &  1.37 &  1.37 &  1.47 &  1.47 &  1.38 \\
C-ResNet50          &  1.78 &  1.69 &  1.68 &   2.42 &   2.42 &  1.86 &  1.48 &  1.48 &  2.20 &  1.90 &  1.47 \\
C-Den.Net121       &  1.86 &  1.62 &  1.60 &   2.77 &   2.82 &  1.81 &  1.42 &  1.40 &  2.00 &  1.96 &  1.40 \\
C-Mob.Netv2       &  1.66 &  1.64 &  1.62 &   2.86 &   2.88 &  1.93 &  1.47 &  1.49 &  2.08 &  2.00 &  1.48 \\
\midrule
I-ResNet50       &  2.81 &  2.65 &  2.67 &   8.00 &   7.97 &  2.67 &  2.65 &  2.64 &  2.92 &  2.93 &  2.65 \\
I-ResNet152      &  2.49 &  2.33 &  2.34 &   7.22 &   7.18 &  2.35 &  2.32 &  2.33 &  2.67 &  2.71 &  2.33 \\
I-VGG19          &  2.94 &  2.92 &  2.92 &   8.64 &   8.63 &  2.94 &  2.97 &  2.94 &  3.21 &  3.15 &  2.94 \\
I-Den.Net169    &  2.48 &  2.35 &  2.35 &   7.33 &   7.31 &  2.37 &  2.34 &  2.34 &  2.74 &  2.80 &  2.35 \\
I-Eff.NetB7 &  2.51 &  2.54 &  2.51 &   6.98 &   6.98 &  2.53 &  2.51 &  2.51 &  2.89 &  2.89 &  2.45 \\
I-Xception       &  2.90 &  2.55 &  2.56 &   7.26 &   7.09 &  2.57 &  2.55 &  2.58 &  2.93 &  3.08 &  2.61 \\
I-Mob.Netv2    &  3.45 &  3.58 &  3.51 &  10.3 &  10.2 &  3.67 &  3.52 &  3.48 &  3.66 &  3.62 &  3.51 \\
\bottomrule
\end{tabular}
    \label{tab:nll}
\end{table}

\begin{table}[h]
    \caption{Brier score for all baselines for CIFAR-10 and Imagenet}

    \centering
\begin{tabular}{lrrrrrrrrrrr}
\toprule
{} &      Base &        TS &       ETS &     TS-IR &        IR &      IRM  &      TS-P &     ETS-P &   TS-IR-P &     IR-P &   IRM-P  \\
\midrule
C-VGG19             &  .731 &  .603 &  .600 &  .617 &  .620 &  .610 &  .565 &  .566 &  .574 &  .574 &  .566 \\
C-ResNet50          &  .677 &  .663 &  .660 &  .681 &  .681 &  .664 &  .622 &  .624 &  .664 &  .659 &  .620 \\
C-Den.Net121       &  .631 &  .600 &  .598 &  .618 &  .618 &  .601 &  .593 &  .587 &  .623 &  .607 &  .584 \\
C-Mob.NetV2       &  .644 &  .640 &  .636 &  .656 &  .656 &  .639 &  .619 &  .627 &  .655 &  .642 &  .621 \\
\midrule
I-ResNet50       &  .667 &  .644 &  .648 &  .692 &  .692 &  .649 &  .646 &  .644 &  .645 &  .643 &  .644 \\
I-ResNet152      &  .620 &  .597 &  .598 &  .645 &  .643 &  .600 &  .597 &  .596 &  .597 &  .597 &  .598 \\
I-VGG19          &  .688 &  .686 &  .687 &  .732 &  .732 &  .687 &  .699 &  .694 &  .691 &  .681 &  .688 \\
I-Den.Net169    &  .620 &  .602 &  .602 &  .650 &  .650 &  .604 &  .600 &  .600 &  .595 &  .596 &  .603 \\
I-Eff.NetB7 &  .621 &  .634 &  .619 &  .635 &  .635 &  .608 &  .617 &  .612 &  .584 &  .586 &  .608 \\
I-Xception       &  .682 &  .625 &  .621 &  .657 &  .661 &  .627 &  .624 &  .620 &  .611 &  .627 &  .635 \\
I-Mob.NetV2    &  .745 &  .767 &  .758 &  .803 &  .802 &  .754 &  .759 &  .751 &  .740 &  .734 &  .750 \\
\bottomrule
\end{tabular}

    \label{tab:brier}
\end{table}

\newpage
To further illustrate the benefit of our modelling approach for different post-calibration methods, we computed for each algorithm the difference in mean ECE between our approach (using a perturbed validation set) and the standard approach (using the unperturbed validation set). Table \ref{tab:deltaECE} highlights that our approach is beneficial for all post-calibration algorithms. 

\begin{table}
    \caption{$\Delta$ECE reveals that using a perturbed validation set for training improves performance across all methods for CIFAR-10 (higher is better).}

    \centering
\begin{tabular}{lrrrrr}
\toprule
{} &  $\Delta$ TS &  $\Delta$ ETS &  $\Delta$ TS-IR &  $\Delta$ IR &  $\Delta$ IRM \\
\midrule
CIFAR VGG19             &  0.661 &   0.622 &     0.706 &  0.718 &   \textbf{0.736} \\
CIFAR ResNet50          &  0.528 &   0.473 &     0.518 &  0.509 &   \textbf{0.575} \\
CIFAR DenseNet121       &  0.103 &   0.158 &     0.376 &  \textbf{0.472} &   0.208 \\
CIFAR MobileNetv2       &  0.281 &   0.113 &     0.428 &  \textbf{0.519} &   0.266 \\
\midrule
ImgNet ResNet50       & -0.022 &   0.365 &     \textbf{0.753} &  0.740 &   0.428 \\
ImgNet ResNet152      &  0.147 &   0.301 &     \textbf{0.778} &  0.708 &   0.276 \\
ImgNet VGG19          & -1.044 &  -0.567 &     0.467 &  \textbf{0.762} &   0.085 \\
ImgNet Den.Net169    &  0.453 &   0.421 &     \textbf{0.795} &  0.662 &   0.118 \\
ImgNet Eff.NetB7 &  0.451 &   0.440 &     \textbf{0.705} &  0.622 &   0.218 \\
ImgNet Xception       &  0.110 &   0.253 &     \textbf{0.715} &  0.221 &  -0.313 \\
ImgNet MobileNetv2    &  0.304 &   0.348 &     0.644 &  \textbf{0.745} &   0.356 \\
\bottomrule
\end{tabular}
    \label{tab:deltaECE}
\end{table}
\newpage
\subsection{Additional experiments}
\begin{figure*}[hbt!]
	\centering
	\begin{subfigure}[h]{0.45\textwidth}
		\centering
    	\includegraphics[width=\textwidth]{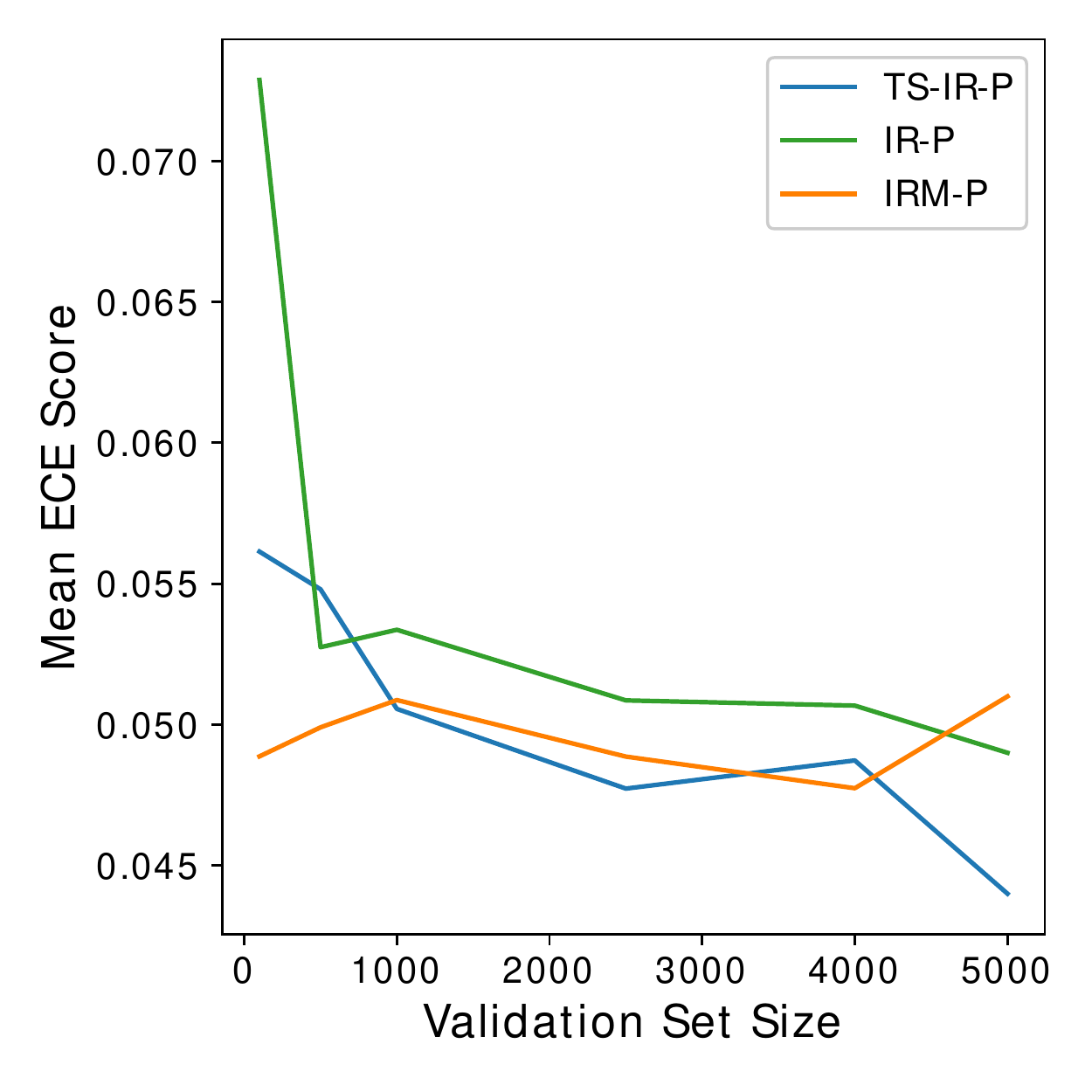}
		\caption{Mean ECE w.r.t. size of validation set}
	\end{subfigure}
	\begin{subfigure}[h]{0.45\textwidth}
		\centering
		\includegraphics[width=\textwidth]{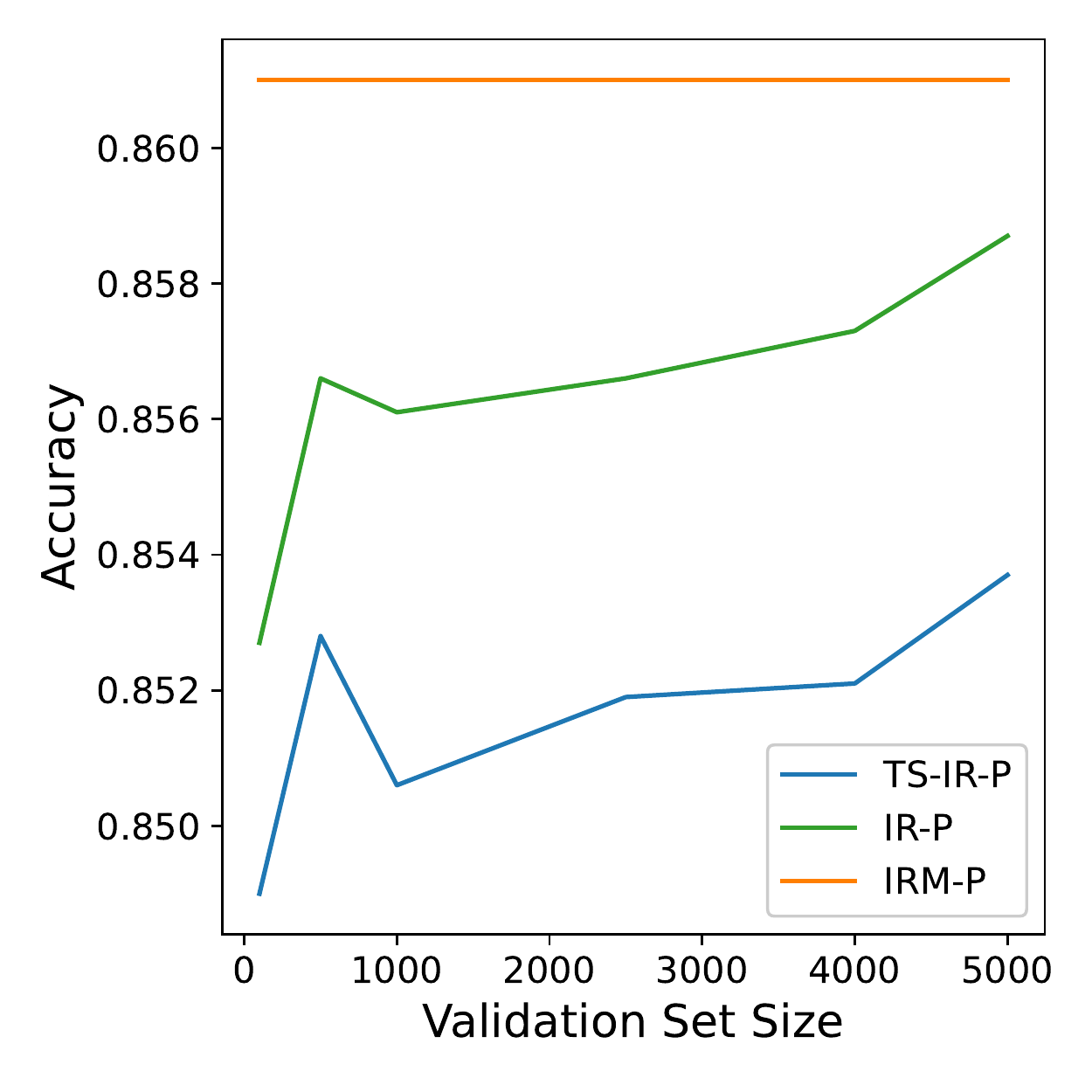}
		\caption{Accuracy w.r.t. size of validation set}
	\end{subfigure}

	\caption{Effect of the chosen size of the validation set on the mean expected calibration error and accuracy scores (CIFAR-10).}
\label{fig:val_set_size}
\end{figure*}

\paragraph{Size of validation set} While both IRM-P and IR-P performed consistently well across baselines, a key difference is that IR-P is not accuracy preserving. In contrast, a model's accuracy remains unchanged after post-calibration with IRM-P. In the main paper, we show that the effect on the accuracy for IR-P is only marginal. To further investigate the robustness of IR-P in terms of accuracy, we assessed the effect of the size of the validation set on performance. Our results show, that in fact for small validation sets accuracy can substantially decrease for IR-P (Fig. \ref{fig:val_set_size} (b)). However, with increasing size of the validation set accuracy increases and ECE decreases (Fig. \ref{fig:val_set_size} (a)). This suggests that for sufficiently large validation set, IR-based methods benefit from their high expressiveness. \\

\paragraph{Type of validation perturbation} Finally, we investigated the effect of the perturbation strategy used to generate a perturbed validation set. To this end, we assessed whether perturbing the validation set using image perturbations rather than the generic perturbations proposed in our work, could lead to similar results. To test this hypothesis, we used the validation perturbations \emph{speckle noise}, \emph{gaussian blur}, \emph{spatter} and \emph{saturate} introduced in \cite{hendrycks2019benchmarking} to generate a perturbed validation set. We then tuned all baselines on this validation set using a VGG19 model trained on CIFAR-10. Table \ref{tab:valH} shows that this resulted in consistently worse calibration errors compared to the generic perturbation strategy proposed in the main paper. This suggests, that our algorithm can indeed yield a validation set that is representative of generic domain drift scenarios.

\begin{table}
    \centering
\begin{tabular}{lrrrrrrrrrrr}
\toprule
Base &        TS &       ETS &     TS-IR   &        IR &     IRM &      TS-H &     ETS-H &   TS-IR-H   &      IR-H &   IRM-H \\

\midrule
 0.323 & 0.158 &  0.152 &  0.173 &  0.176 &  0.167 & 0.102 & \textbf{0.096} & 0.112 &  0.127 &  0.114\\
\bottomrule
\end{tabular}
\caption{Mean expected calibration error across all test domain drift scenarios (affine transformations for CIFAR-10). Tuning was performed on the validation set and the perturbed validation set generated by applying the validation perturbations proposed in \cite{hendrycks2019benchmarking}. The latter is denoted by the suffix -H.}\label{tab:valH}
\end{table}

\section{Additional Guidelines}
Based on our extensive experiments, we propose the following guidelines for practitioners:

\begin{itemize}
    \item If a sufficiently large validation set is available and calibration for in-domain settings is of particular concern, we recommend using IR-P or TS-IR-P. This may result in changes in model accuracy.
    \item If the practitioner requires that the accuracy of the trained model remains unchanged or truly OOD scenarios are of particular concern, we recommend using IRM-P or ETS-P.
\end{itemize}



\end{document}